\documentclass[10pt,journal,compsoc]{IEEEtran}
\usepackage{amsmath,amsfonts}
\usepackage{multirow}
\usepackage{ragged2e}
\usepackage{array}
\usepackage[caption=false,font=normalsize,labelfont=sf,textfont=sf]{subfig}
\usepackage{textcomp}
\usepackage{stfloats}
\usepackage{url}
\usepackage{verbatim}
\usepackage{graphicx}
\usepackage{hyperref}
\usepackage{cite}
\usepackage{color}
\usepackage{CJKutf8}
\usepackage[switch]{lineno}
\newcommand{\ch}[1]{\begin{CJK}{UTF8}{gbsn}{#1}\end{CJK}}
\usepackage[ruled,vlined]{algorithm2e}
\definecolor{commentcolor}{RGB}{110,154,155}   % define comment color
\newcommand{\PyComment}[1]{\ttfamily\textcolor{commentcolor}{\# #1}}  % add a "#" before the input text "#1"
\newcommand{\PyCode}[1]{\ttfamily\textcolor{black}{#1}} % \ttfamily is the code font
\usepackage{listings}
\begin{document}

\title{Instruction-Guided Scene Text Recognition}


\author{Yongkun~Du,
        Zhineng~Chen,\IEEEmembership{~Member,~IEEE,}
        Yuchen~Su,
        Caiyan~Jia,
        and~Yu-Gang~Jiang,~\IEEEmembership{Fellow,~IEEE}
\thanks{This work was supported by the National Key R\&D Program of China (2022YFB3104703), and in part by the National Natural Science Foundation of China (32341012).}
\IEEEcompsocitemizethanks{
\IEEEcompsocthanksitem{Yongkun Du, Zhineng Chen, Yuchen Su and Yu-Gang Jiang are with the School of Computer Science, Fudan University, Shanghai 200433, China.
E-mail: \{ykdu23, ycsu23\}@m.fudan.edu.cn, \{zhinchen, ygj\}@fudan.edu.cn.}
\IEEEcompsocthanksitem{Caiyan Jia is with the School of Computer and Information Technology, Beijing Jiaotong University, Beijing 100044, China.
E-mail: cyjia@bjtu.edu.cn.}
}
\thanks{Corresponding author: Zhineng~Chen.}% <-this % stops an unwanted space
%\thanks{Manuscript received September, 2023.}}
}


\markboth{For review only}%
{Shell \MakeLowercase{\textit{et al.}}: A Sample Article Using IEEEtran.cls for IEEE Journals}


\IEEEtitleabstractindextext{%
\begin{abstract}

Multi-modal models have shown appealing performance in visual recognition tasks, as free-form text-guided training evokes the ability to understand fine-grained visual content. However, current models cannot be trivially applied to scene text recognition (STR) due to the compositional difference between natural and text images. We propose a novel instruction-guided scene text recognition (IGTR) paradigm that formulates STR as an instruction learning problem and understands text images by predicting character attributes, e.g., character frequency, position, etc. IGTR first devises $\left \langle condition,question,answer\right \rangle$ instruction triplets, providing rich and diverse descriptions of character attributes. To effectively learn these attributes through question-answering, IGTR develops a lightweight instruction encoder, a cross-modal feature fusion module and a multi-task answer head, which guides nuanced text image understanding. Furthermore, IGTR realizes different recognition pipelines simply by using different instructions, enabling a character-understanding-based text reasoning paradigm that differs from current methods considerably. Experiments on English and Chinese benchmarks show that IGTR outperforms existing models by significant margins, while maintaining a small model size and fast inference speed. Moreover, by adjusting the sampling of instructions, IGTR offers an elegant way to tackle the recognition of rarely appearing and morphologically similar characters, which were previous challenges. Code: \url{https://github.com/Topdu/OpenOCR}.

\end{abstract}

}
\maketitle
\IEEEdisplaynontitleabstractindextext

\section{Introduction}

\IEEEPARstart{S}{cene} text recognition (STR) is a longstanding pattern recognition task that focuses on reading natural text images. Essentially, STR is a multi-modal task that learns a mapping from image modality to text modality, aiming to decipher the character sequence. Over the past years, a magnitude of methods have been devoted to STR and different recognition pipelines like parallel recognition (PR) \cite{shi2017crnn,yu2020srn,fang2021abinet,mgpstr,duijcai2022svtr,du2023cppd,ijcai2023LPV} and auto-regressive recognition (AR) \cite{shi2019aster,li2019sar,Sheng2019nrtr,BautistaA22PARSeq} have been developed, which satisfy the diverse accuracy and speed needs in various applications.

Recently, there is a trend to develop multi-modal models \cite{li2022glip, liu2023grounding, Kirillov_2023_sam, Wang_2023_ICCVseggpt} for the generic visual recognition tasks \cite{long2015fully, girshick2014rich}. 
Notably, pioneering models such as Grounding-DINO \cite{liu2023grounding} and SAM \cite{Kirillov_2023_sam} have embraced the integration of natural language as instructive guidance, which enables a profound understanding of fine-grained visual content. For instance, by using free-form text rather than categorical labels for training \cite{liu2023grounding}, the learned model is capable of understanding more specific objects like \emph{the left lion} rather than \emph{lion}, and meanwhile, resulting in superior performance compared to uni-modal models in typical benchmarks \cite{2014eccvmscoco,imagenet1k}.

\begin{figure}
  \centering
\includegraphics[width=0.48\textwidth]{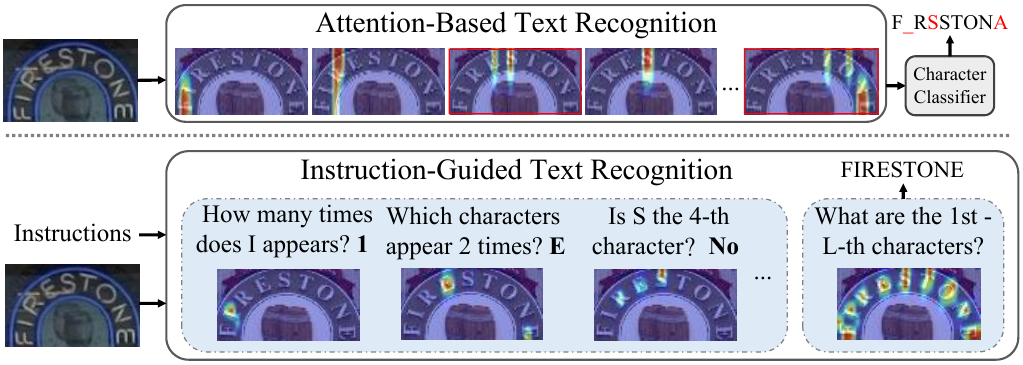} 
  \caption{\textbf{Upper}: popular attention-based STR models follow the pipeline of positioning visual features for every character and then classifying. Mis-recognitions may happen if features are positioned incorrectly. \textbf{Bottom}: IGTR comprehends the text image first and then recognition. It understands the question and associates its answer with the corresponding visual features, generating robust instruction-guided STR.}
  \label{fig:fig1}
\end{figure}

Similarly, STR models could benefit significantly from an enhanced understanding of text images. However, given the distinct compositional elements of natural and text images, applying existing instruction learning schemes \cite{pmlr-v139-clip,liu2023grounding,Kirillov_2023_sam,liu2023visualllava,liu2024textmonkey} to this context poses substantial challenges. Unlike natural images which primarily depict objects and scenes from the physical world and can be effectively described using natural language, text images typically represent single words. Treating a word simply as the language provides limited semantic context to guide STR models in understanding the text images. Therefore, developing dedicated instructional strategies for STR tasks is an urgent necessity. Note that the text image is composed of sequentially arranged characters, necessitating the exploration of character-central properties to facilitate a deep understanding of the text content. However, the exploration of these properties and their contributions to text recognition has received limited attention in existing solutions. Furthermore, with the increasing use of Optical Character Recognition (OCR) models in mobile and edge computing scenarios, where lightweight STR models \cite{shi2017crnn,duijcai2022svtr,ppocrv3} are required, it is yet to explore further multi-modal STR models that run fast and are less demanding in computational resources.

In this paper, we aim to build an efficient instruction-learning-based paradigm for STR. Our proposed method, termed instruction-guided STR (IGTR), firstly attempts to make meaningful efforts in both instruction construction and architecture design. In terms of instruction construction, we define character attributes as properties regarding the status, frequency, and position of one or multiple characters within text. We argue that character attributes are critical components to establishing a deep understanding of text. Then, we develop two types of instructions customized for character attribute prediction and text recognition. For character attribute prediction, we devise rich and diverse instructions in the form of $\left \langle condition,question,answer\right \rangle$, as shown in Tab. \ref{tab:instruction}. The \emph{condition} denotes known attributes, and \emph{question-answer} is question-answer pairs regarding attributes or attribute combinations. They represent learning data and labels for model training, aiming to evoke a human-like understanding of text images. Meanwhile, recognition instructions of the same form are proposed to consolidate the model training and perform text recognition. They simulate different text recognition pipelines, including PR, AR, and other useful but less-explored ones, as shown in Tab. \ref{tab:rec_instruction}.

\begin{table*}[t]\footnotesize
\centering
\caption{Character attribute prediction instructions (taking \emph{ARTETA} as the example), where \textcolor{red}{red} in \emph{Question} column denotes character or numerical variables. Different colors in \emph{Type} column represent different types of questions. Character is abbreviated to char. The same below.}
\setlength{\tabcolsep}{2pt}{
\begin{tabular}{l|l|l|l|l}
    \hline
Condition                                                                  & \begin{tabular}[c]{@{}c@{}} Condition\\ Option \end{tabular} & Question                                                     & Answer           &     Type                              \\
    \hline
\multirow{12}{*}{\begin{tabular}[c]{@{}l@{}}1. Char \{$c_{i}$\} in the image\\ 2. Char \{$c_{i}$\} appear \{$N_i$\} times\\ 3. The $i$-th char are $c_{i}$\\ 4. Sub-string \{$c_i$ - $c_{i+l_s}$\} in the image\\ 5. There are L chars in the image\\ 6. None\end{tabular}}                               & 1/2/3/4/5/6   & How many \textcolor{red}{A} in the image?                                       & 2                & \textcolor{red}{Frequency}                    \\
& 1/2/3/4/5/6   & How many \textcolor{red}{A} in the first \textcolor{red}{3} chars?                 & 1                & Constrainted \textcolor{red}{frequency}              \\
       & 1/2/3/4/5/6   & Does \textcolor{red}{A} appear \textcolor{red}{2}/\textcolor{red}{1} times in the image?                                       & Yes/No           & \textcolor{yellow}{Status}          \\
& 1/2/3/4/5/6   & Does \textcolor{red}{A} appear \textcolor{red}{2}/\textcolor{red}{1} times in the first \textcolor{red}{3} chars?              & No/Yes           & Constrainted \textcolor{yellow}{status}              \\
  & 1/2/3/4/5/6   & Which char in the image is \textcolor{red}{E}?                                              & 4-th & \textcolor{green}{Position}                   \\                                                          & 1/2/3/4/5/6   & Is \textcolor{red}{A} the \textcolor{red}{1}st/\textcolor{red}{5}-th char in the image?                                     & Yes/No          & Search \textcolor{yellow}{status}     \\
  & 1/2/3/4/5/6   & Which chars appear \textcolor{red}{2} times?                        & A and T          & \textcolor{blue}{Char}  \\
     & 1/2/3/4/5/6   &  Which chars appear \textcolor{red}{1} time in first \textcolor{red}{3} chars? & A, R and E       & Constrainted \textcolor{blue}{char} \\
& 1/2/3/4/5/6     & Where is the sub-string \textcolor{red}{RTE}/\textcolor{red}{RTS}?              & 2nd/None                & Sub-string \textcolor{green}{position}                 \\
& 1/2/3/4/5/6     & Is sub-string \textcolor{red}{RTE} at position \textcolor{red}{2}/\textcolor{red}{3}?              & Yes/No                & Sub-string \textcolor{yellow}{status}                  \\
& 1/2/3/4/6     & How many chars in the image?              & 6                & Length \textcolor{red}{frequency}                 \\

& 5/6           & What are the first and last chars?               & A/A              & Edge \textcolor{blue}{char}         \\
                 \hline
\end{tabular}}

\label{tab:instruction}
\end{table*}

\begin{table*}[t]\footnotesize
\centering
\caption{Text recognition instructions, where each row simulates a different recognition pipeline.}
\setlength{\tabcolsep}{2pt}{
\begin{tabular}{l|l|l|l}
\hline
Condition & Question                                                     & Answer           &     Type                              \\
\hline
None                                                                                 & What are the $1$st to L-th chars in the image?    & $c_1$ - $c_{L}$                & Parallel Recognition (PR)                           \\
$1$st to $i$-th chars are $c_1$ - $c_{i}$                                   & What is the $i$+$1$-th char? & $c_{i+1}$ & Auto-regressive Recognition (AR) \\

$1$st to $i$-$1$-th, $i$+$1$-th to $j$-th chars are known &   What is the $i$-th char?      & $c_{i}$   & Re-Identification (RI)                    \\
There is a sub-string $c_i$ - $c_{i+l_s}$                                       & What is the previous/next char of the sub-string?                 & $c_{i-1}$/$c_{i+l_s+1}$            & Extrapolating Recognition (ER)                     \\
\hline
\end{tabular}}

\label{tab:rec_instruction}
\end{table*}

With the instructions, we develop a dedicated architecture for both attribute prediction and text recognition. It consists of a lightweight instruction encoder to generate textual embeddings based on instructions, a cross-modal feature fusion module for characterizing interactions between image and text modalities, and a multi-task answer head responsible for answering different questions and reading the text. IGTR is trained using a large number of attribute prediction and recognition instructions, which endows the model with a profound understanding of character attributes. Then, text images can be accurately recognized with the distinct recognition pipeline in Tab. \ref{tab:rec_instruction}.

We conduct extensive experiments on English and Chinese benchmarks to analyze the performance of IGTR. No matter following PR or AR pipeline, IGTR outperforms existing models by clear margins in terms of accuracy, while maintaining a small model size and fast inference speed. Meanwhile, the attribute prediction instructions guide meaningful text understanding, even when solely trained on them, IGTR exhibits impressive recognition capability. Furthermore, IGTR benefits from remarkable learning flexibility inherited from its instruction-guided nature. By simply adjusting the rule of instruction sampling, IGTR exhibits superior performance in recognizing both rarely appearing and morphologically similar characters, which have been persistent challenges for previous methods.

IGTR holds the potential to represent a novel paradigm of STR. As exemplified in Fig. \ref{fig:fig1}, the top half represents the recognition pipeline of popular attention-based STR models. They utilize the attention mechanism to position visual features for every character and then use the features to perform character classification. Mis-recognitions may happen if features are positioned incorrectly. In contrast, as shown in the bottom half, IGTR successfully grasps character attributes by exhaustive question-answering-based learning, and then comprehends the text image. Therefore visual features associate characters correctly and IGTR generates more robust recognition. On the other hand, recent Multi-modal Large Language Models (MLLMs) such as LLaVA \cite{liu2023visualllava} and Monkeys \cite{li2023monkey,liu2024textmonkey}, trained or fine-tuned on OCR-relevant dialog datasets \cite{singh2021textocr}, also show remarkable performance in multiple OCR-related tasks including STR. However, these models typically use large language models to compile the text modality, thus having huge parameters and being highly demanding in computational resources. In contrast, the model size of IGTR is 24.1M, and different IGTR models consume 4 ms-10 ms only when inferring a text instance in one NVIDIA 1080Ti GPU, both are appealing properties in mobile and edge computing applications.

Contributions of this paper are summarized as follows.
\begin{itemize}
  \item We propose IGTR that regards STR as a cross-modal instruction learning task. Unlike existing models, IGTR facilitates text recognition by comprehending text attributes, presenting a novel paradigm for STR.
  \item We introduce rich and diverse triplet-form instructions $\left \langle condition,question,answer\right \rangle$, and develop a dedicated architecture to effectively learn these instructions and the text image, establishing the first instruction-guided STR solution. 
  \item Through extensive English and Chinese experiments, we demonstrate the effectiveness of IGTR not only in public benchmarks, but also in offering a versatile framework to tackle prevalent STR challenges.
\end{itemize}

\section{Related Work}

\subsection{Scene Text Recognition (STR)} 
We can broadly classify existing sequence-based STR models into two types according to their recognition pipelines, i.e., PR and AR. PR identifies all the characters within a text image at once. It can be further classified as Connectionist Temporal Classification (CTC)-based and parallel decoding ones. The former \cite{shi2017crnn, hu2020gtc, duijcai2022svtr} assumes that the text image can be implicitly split into a series of stripes from left to right, each corresponding to a recognition unit, i.e., a character or blank. The recognized unit sequence is then optimized by the CTC rule \cite{CTC} to get recognition results. The second \cite{yu2020srn, qiao2021pimnet, fang2021abinet, Wang_2021_visionlan, du2023cppd, ijcai2023LPV} commonly adopt an attention-based encoder-decoder framework. They first allocate a fixed number of placeholders and then learn to fill in the placeholders with proper features and deduce the characters in parallel. To compensate for semantic context, recent studies propose various ways, for example, knowledge distillation \cite{qiao2021pimnet}, external language models \cite{fang2021abinet, Wang_2021_visionlan, ijcai2023LPV}, character counting \cite{du2023cppd}, etc., which improve the recognition accuracy remarkably.

In contrast, AR \cite{shi2019aster, li2019sar, Sheng2019nrtr, lee2020recognizing} adopts a one-by-one recognition pipeline. These models leverage not only visual features but also previously decoded characters as language clues to enhance recognition accuracy, typically yielding superior results. However, they may encounter challenges such as attention drift when faced with rarely seen patterns like highly deformed text \cite{yue2020robustscanner}. To mitigate attention drift, some methods propose to incorporate the position clue into the recognition process \cite{wan2020textscanner,yue2020robustscanner,zheng2024cdistnet}. Additionally, methods like permuted and bidirectional recognition are developed \cite{BautistaA22PARSeq,bleeker2019bidirectional}, providing a stronger language prior. Generally, AR models provide more natural and finer language integration compared to PR models. However, their sequential decoding nature results in slower inference speed, limiting their use in time-sensitive applications. 

In addition to the two types above, there are studies employing double-check-like techniques to assist recognition. For example, some methods incorporate character re-identification strategies \cite{yu2020srn,fang2021abinet,BautistaA22PARSeq,MATRN}. Typically, they use language models to rectify initial outputs generated by PR or AR models. Despite improving the accuracy, they also incur a possibility of introducing errors, especially when dealing with contextless text.

\subsection{Multi-Modal Visual Recognition} 
Multi-modal models \cite{long2015fully, girshick2014rich} have achieved promising performance recently, particularly in visual recognition tasks, as instruction learning can evoke an appealing understanding of fine-grained visual content. CLIP \cite{pmlr-v139-clip} and ALIGN \cite{jia2021align} are pioneering studies in this direction. They have performed cross-modal contrastive learning on billion-level image-text pairs, enabling impressive zero-short image classification \cite{imagenet1k}. Later, GLIP \cite{li2022glip} and Grounding-DINO \cite{liu2023grounding} introduce instruction-based object detection \cite{2014eccvmscoco}. They utilize instructions of free-form language for training, and can detect objects that exactly match the language. Similar approaches are explored in image segmentation, for instance, the well-regarded SAM \cite{Kirillov_2023_sam} and SegGPT \cite{Wang_2023_ICCVseggpt}. These models leverage rich interactive instructions including text to achieve generic and deep visual understanding, enabling segmentation of anything. Generally, these studies mainly focused on more generic visual tasks which involve multi-object, multi-scene contexts. However, text images primarily focus on character-central information. This distinction requires a specific adaptation of the instruction-guided approach to capture the fine-grained attributes unique to text, which is seldom discussed before. Note that instruction-based recognition has also been applied to tasks like license plate recognition \cite{2023VQACLPR} and speech recognition \cite{lai2023instrspeech}. However, they are more like the application of VQA models \cite{wang2022ofa, antol2015vqa} or generative models \cite{zhang2023speechgpt, deshmukh2023pengi} to these tasks.

Note that recent multi-modal large language models (MLLMs) \cite{zhu2023minigpt4,liu2023visualllava,li2023monkey,liu2024textmonkey} have also shown impressive performance on OCR tasks. These models typically implement OCR in two ways. One branch is prompt engineering on pre-trained visual-language models \cite{gu2023systematic}, which performs recognition based solely on prompts without updating model parameters. Typically, this approach involves querying the model with an image and a text question to generate responses \cite{alayrac2022flamingo,zhu2023minigpt4}, which can extend to handling OCR-related queries. Another branch focuses on fine-tuning MLLMs on OCR data, for example, LLaVA \cite{liu2023visualllava} and Monkeys \cite{li2023monkey,liu2024textmonkey}. These models demonstrate capabilities in various OCR-related tasks like document understanding and key information extraction without relying on traditional OCR tools \cite{ppocrv3}. Nevertheless, the two branches above typically incorporate large language models to better capture textual information. As a result, they consume substantial resources, restricting their deployment in mobile and edge computing scenarios where lightweight STR models \cite{shi2017crnn,duijcai2022svtr,ppocrv3,zheng2023TPS++} are prevalent. In contrast, IGTR simplifies textual query modeling into a straightforward embedding format, eliminating the need for large language models and significantly reducing the model's computational overhead.

\section{Method}
\label{sec:Method}

\begin{figure*}
  \centering
\includegraphics[width=0.98\textwidth]{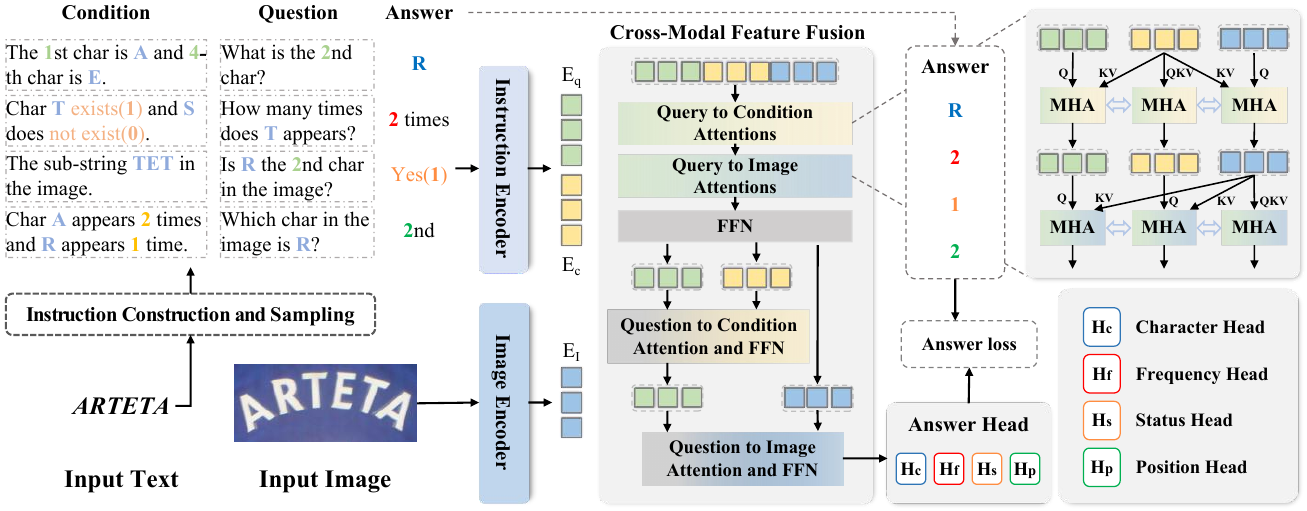} 
  \caption{Overview of Instruction-Guided STR. Instruction triplets $\left \langle condition,question,answer\right \rangle$ are sampled from text \emph{ARTETA}, where \emph{condition} and \emph{question} are encoded as the corresponding embeddings by the instruction encoder. Meanwhile, the image embedding is extracted from the image encoder. The three embeddings are interacted and fused by the cross-modal feature fusion module, and a multi-task answer head is appended to answer different types of questions. The whole architecture enjoys a lightweight design and ent-to-end optimization. Best viewed in color.}
  \label{fig:IGTR_overview}
\end{figure*}

Our method consists of two parts, i.e., instruction construction and sampling, IGTR architecture and learning. The instructions, including both character attributes prediction and recognition ones, are $\left \langle condition,question,answer\right \rangle$ triplets. The triplet composition ensures that many instructions can be generated for attribute learning even for a short text, and meanwhile, different recognition pipelines can be precisely described. On the other hand, IGTR architecture mainly consists of a lightweight instruction encoder, a cross-modal feature fusion module and a multi-task answer head. They together enable efficient encoding of the instructions, and effective cross-modal learning and recognition.

\subsection{Instruction Construction and Sampling}

 We have defined a series of character attribute prediction and text recognition instructions. Each instruction consists of a \emph{condition}, a \emph{question}, and an \emph{answer}. The \emph{condition} means knowledge given in advance, and the \emph{question}-\emph{answer} pair corresponds to the specified query and its response. In Tab. \ref{tab:instruction}, \emph{Condition Option} describes the compatibility between conditions and question-answers for character attribute prediction instructions. These questions, together with the corresponding conditions, compel the model to assess the character from different aspects, determining whether it contributes to the answer. This process, in turn, guides the model towards nuanced text understanding. Note that the question-answers in Tab. \ref{tab:instruction} can be classified into four types according to their answers (highlighted by different colors in \emph{Type} column), i.e., \emph{character}, \emph{frequency}, \emph{position}, and \emph{status}.

We further define five types of character attributes: character-status (\texttt{cs}) represents whether a specific character exists, character-frequency (\texttt{cf}) denotes the occurrence times of a specific character, constrained character-frequency (\texttt{cf}$_{cons}$) describes character-frequency given a specific constraint (e.g., the first 3 characters), position-character (\texttt{pc}) means the character at the \emph{i}-th position, and sub-string (\texttt{ss}) denotes a certain length sub-string. For example, given text \emph{ARTETA}, assuming the constraint is \emph{the first 3 characters} and the sub-string length $l_s$ is set to 3, character attributes can be obtained by traversing the text, as expressed in the following set form:
\begin{align}
    &\texttt{cs} = \text{[[A,1], [B,0], ..., [E,1], ..., [R,1], [S,0], [T,1], ..., [Z,0]]} \notag \\
    &\texttt{cf} = \text{[[A,2], [E,1], [R,1], [T,2]]} \notag \\
    &\texttt{cf}_{cons=3} = \text{[[A,1], [E,0], [R,1], [T,1]]} \notag \\ 
    &\texttt{pc} = \text{[[0,A], [1,R], [2,T], [3,E], [4,T], [5,A]]} \notag \\
    &\texttt{ss}_{l_s=3} = \text{[[0, ART], [1, RTE], [2, TET], [3, ETA]]} \notag 
\end{align}

Note that each element in those sets is represented by a square bracket with two units, corresponding to the variables in question and answer, respectively.

We propose to enlarge the traditional question-answer instruction to $\left \langle condition,question,answer\right \rangle$ triplet. Specifically, a partitioning strategy is proposed to split character attributes into two parts: one is \emph{condition} to represent already known attributes, and the other is attributes left for question-answer-based model learning. Assuming that the \emph{condition} is [[E,1], [R,1], [S,0]], [[E,1],[T,2]] and [[0,A], [2,T], [5,A]], which are subsets of \texttt{ce}, \texttt{cf} and \texttt{pc}, respectively. The \emph{condition} can be described as follows: \emph{``Characters E and R exist in the image, but S does not exist''}; \emph{``Character T appears twice and E appears once in the image''}; and \emph{``The 1st/3rd/6-th character in the image is A/T/A''}. Take the set split on \texttt{pc} as an example, the remaining subset is [[1,R], [3,E], [4,T]], allowing questions such as \emph{``Which character in the image is $R$/$E$/$T$?''} with answers \emph{``the 2nd/4-th/5-th one''} for model training. Other question-answer pairs are generated similarly. According to the types of answer, questions and answers are categorized into four types, i.e., \texttt{Ques} = [$Q_c$,$Q_f$,$Q_p$,$Q_s$] and \texttt{Ans} = [$A_c$,$A_f$,$A_p$,$A_s$], where the suffixes $c$, $f$, $p$, and $s$ denote the answer type is \emph{character}, \emph{frequency}, \emph{position}, and \emph{status}, respectively.

The triplet form notably increases the richness and diversity of instructions. To sample instructions for each input text during model training, we repeat the partitioning strategy (\texttt{GenInsts}) \emph{K} times, with the \emph{condition} randomly shuffling and splitting each time to ensure disjoint and varied instructions. This process is carried out by \texttt{RandomSplit(Attributes)}, as described in Algorithm \ref{algo:igtr}. For each partition, we enumerate all possible question-answer variables and up to $\sum\limits_{p = 1}^N {{(L-p)}\frac{N!}{p!(N-p)!}}$ different instructions could be sampled, where $N$ is the number of attributes \texttt{len(atrb)}, \texttt{p} represents the number of attributes in \emph{condition} and it is randomly obtained using \texttt{random.randint(len(atrb))}. By sampling \emph{K} times, we can get an even larger number of instructions.

\begin{algorithm}[!htbp]\footnotesize

\caption{Pseudo-code of instruction sampling and model optimization in IGTR}
\label{algo:igtr}
\PyCode{import random} \\
\PyComment{text, image from training dataset} \\
\PyComment{cs: character-status attribute set} \\
\PyComment{cf: character-frequency attribute set} \\
\PyComment{pc: position-character attribute set} \\
\PyComment{ss: sub-string attribute set} \\
\PyCode{def RandomSplit(atrbs):} \\
\Indp
    \PyCode{Cond=[], QA=[]} \\
    \PyCode{for atrb in atrbs:} \\
    \Indp
        \PyCode{p = random.randint(len(atrb))} \\
        \PyCode{random.shuffle(atrb)} \\
        \PyCode{Cond.append(atrb[p:])} \\
        \PyCode{QA.append(atrb[:p])} \\
    \Indm
    \PyCode{return Cond, QA}\\
\Indm  
\PyCode{def GenInsts(Attributes):} \\
\Indp
    \PyCode{Cons = random.randint(len(text)} \\
    \PyComment{Generating attributes for constraints} \\
    \PyCode{cf$_{cons}$ = GenConsCF(text, Cons)}\\
    \PyCode{Attributes += [cf$_{cons}$]}\\
    \PyCode{Cond, QA = RandomSplit(Attributes) } \\
    \PyComment{grouping QA into four types by answer} \\ 
    \PyCode{Ques, Ans = GenQAbyType(QA, text)} \\
    \PyCode{return Cond, Ques, Ans} \PyComment{instruction} \\
\Indm  
\PyCode{Attributes = [cs, cf, pc, ss]} \\
\PyCode{$E_I$ = ImageEncoder(image)} \\
\PyCode{LossSum = 0} \\
\PyCode{for k in range(K):} \\
\Indp   % start indent
    \PyCode{Cond, Ques, Ans = GenInsts(Attributes)} \\
    \PyCode{$E_c$, $E_q$ = InstsEncoder(Cond, Ques)}  \\
    \PyCode{AnsRelatedEmbs = CMFF($E_q$, $E_c$, $E_I$)} \\
    \PyCode{AnsPred = AnswerHead(AnsRelatedEmbs)}\\
    \PyCode{LossSum += AnswerLoss(AnsPred, Ans)}\\
\Indm  
\PyCode{LossSum.backward()} \PyComment{back-propagate} \\ 
\PyCode{update(IGTR)} \PyComment{AdamW} \\
\Indm

\end{algorithm}

\subsection{IGTR Architecture and Learning}
The IGTR architecture is depicted in Fig. \ref{fig:IGTR_overview}. We also take \emph{ARTETA} as the example for illustration.
First, the sampled instructions are fed into an instruction encoder to generate \emph{condition} and \emph{question} embeddings. Meanwhile, an image encoder extracts the image embedding from image \emph{ARTETA}. These embeddings are then fed into a cross-modal feature fusion module (CMFF) to absorb answer-related cross-modal features. In the following, the absorbed embeddings are forwarded to a multi-task answer head, which has four heads aligned with the four types of answers, and each head is responsible for one answer type.

\noindent\textbf{Image encoder.} 
On the image side, SVTR-B backbone \cite{duijcai2022svtr} (with the rectification module and CTC decoder removed) is utilized as the encoder. Given a text image of size $H \times W \times 3$, visual features ($F_v \in \mathbb{R}^{\frac{H}{8} \times \frac{W}{4} \times D}$) are extracted by SVTR-B \cite{duijcai2022svtr}. Subsequently, the image embedding ($E_I \in \mathbb{R}^{\frac{HW}{32} \times D}$) is obtained by flattening the height ($H$) and width ($W$) of $F_v$. Here, $D=384$ represents the dimension of the features.

\noindent\textbf{Instruction encoder.}
The instruction encoder aims to extract distinct features for each instruction. Since both \emph{condition} and \emph{question-answer} are composed of character attributes, we can describe the instruction using attribute-level representation. Therefore, unlike previous methods that employ pre-trained text encoders \cite{bert, pmlr-v139-clip} to compile the instruction, which is computationally costly, we devise a lightweight encoder to map the instruction to \emph{condition} and \emph{question} embeddings.

\begin{figure*}
 \centering
\includegraphics[width=0.94\textwidth]{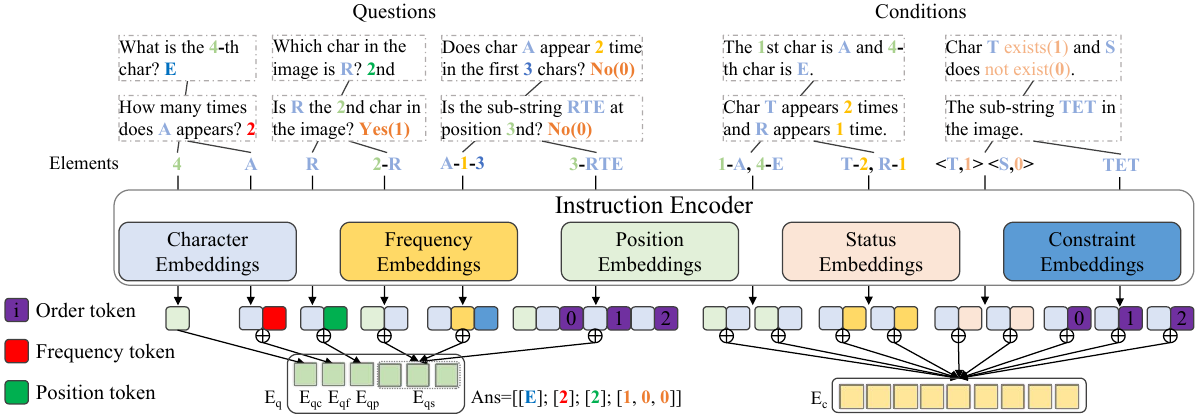} 
 \caption{Details of the instruction encoder. It maps different instructions to feature embeddings by using the five learnable embedding layers placed in the middle of the figure, where different attribute elements or additional tokens are summed to generate a unique \emph{condition} or \emph{question} representation. All \emph{condition} embeddings are concatenated to generate a partition-level representation, and all \emph{question} embeddings are grouped into four classes according to their answer types. Best viewed in color.}
 \label{fig:instruct_encoder}
\end{figure*}

We first inspect the compositional elements of the character attributes, which consist of five basic attribute elements, i.e., character, frequency, position, status, and constraint. All are enumerable. Therefore, as depicted in Fig. \ref{fig:instruct_encoder}, the instruction encoder consists of five learnable embedding layers. They are character embedding ($CE \in \mathbb{R}^{C \times D}$), frequency embedding ($FE \in \mathbb{R}^{F \times D}$), position embedding ($PE \in \mathbb{R}^{P \times D}$), status embedding ($SE \in \mathbb{R}^{2 \times D}$), and constraint embedding ($ConsE \in \mathbb{R}^{L \times D}$) layers. In these embeddings, $C$, $F$, and $P$ signify the size of the character set, the maximum character frequency, and the maximum position index, respectively. We set $F$, $P$ and $L$ to the same value. For each attribute element, the encoding begins by mapping it to the index in the corresponding set. Then, its embedding is found by looking up the vector at this index in the corresponding embedding layer.

We then represent each \emph{condition} and \emph{question} using attribute elements. For those associated with only one attribute element, they are directly represented by the embedding of that element. For instance, \emph{``What is the 4-th character?''} is described by a 384-d embedding denoting 4 from $PE$. Note that \emph{condition} and some \emph{question} are related to multiple attribute elements, e.g., \texttt{ce}, \texttt{cf} and \texttt{pc}. Their representations are obtained by summing the embeddings of its constituent elements. As illustrated in Fig. \ref{fig:instruct_encoder}, \emph{``The 1st character is A''} can be written as \emph{1-A}, and the embedding is the sum of the corresponding vectors from $PE$ and $CE$. Moreover, for questions related to $\texttt{cf}_{cons}$, the embedding is further added with a constraint embedding from $ConsE$. To precisely describe a substring, we first introduce an order token ($ot \in \mathbb{R}^{1 \times D}$) to distinguish the order of characters within a substring. Then, each character in the substring is represented by the sum of its character embedding and corresponding order token, and the substring is represented by further summing the obtained embeddings. In addition, since character position and frequency questions (see Tab. \ref{tab:instruction}) are both described by unitary attribute embedding, a position token ($pt \in \mathbb{R}^{1 \times D}$) and a frequency token ($ft \in \mathbb{R}^{1 \times D}$) are also defined. They are added to the corresponding unitary attribute embedding for differentiation. As the example, \emph{``How many times does A appears?''} can be described by an embedding that sums $ft$ and the corresponding ``A'' vector from $CE$, and \emph{``Which character in the image is R?''} is characterized by an embedding summed $pt$ from the vector denoting ``R'' in $CE$.

With the mapping above, we get the partition-level \emph{condition} and \emph{question} embeddings. The former, i.e., $E_c \in \mathbb{R}^{L_c \times D}$, is obtained by concatenating all $L_c$ \emph{condition} embeddings in that partition. While the latter is obtained by grouping the \emph{question} embeddings into four sets according to answer types, i.e., $E_q$ = [$E_{qc}$;$E_{qf}$;$E_{qp}$;$E_{qs}$] $\in \mathbb{R}^{L_q \times D}$, where $E_{qc}$ corresponds to question set $Q_{c}$ and the rest similarly defined, [;] denotes feature concatenation, $L_q$ is the number of questions. Although a large number of instructions are sampled given a text instance, the \emph{K} partitions and the questions in each partition are independent to each other. They thus can be processed in parallel, which enjoys efficient computation with almost no increase in training cast.

To enable parallel processing, we process the \emph{K} partitions sequentially, predicting the four types of answers independently for each partition. This approach ensures consistent shapes for both \emph{condition} and \emph{question} embeddings across all partitions. Specifically, when the length of $E_c$, $E_{qc}$, or $E_{qs}$ falls short of $P$, they are padded with placeholder tokens to maintain consistent shapes. For $E_{qf}$ and $E_{qp}$, their shapes are both set to \( C \times D \). The padded tokens are subsequently masked during the multi-head attention computation and ignored in loss calculations. As a result, this padding strategy facilitates seamless handling of variable-shape embeddings as inputs and ensures efficient parallel training within Transformer-based architectures.

\noindent\textbf{Cross-modal feature fusion.}
Once the image and instructions are encoded, effectively fusing them is crucial for multi-modal tasks. A common practice involves treating the instruction embedding as the \emph{Query} and the image embedding as the \emph{Key} and \emph{Value}, performing cross-attention to extract relevant information from the image modality. Since the instruction in Tab. \ref{tab:instruction} includes \emph{condition} that represents the known fact, and \emph{question} that denotes the asked question, it is necessary to construct a new module to enable the effective interaction and fusion of these three embeddings.

We devise a dedicated cross-modal feature fusion (CMFF) module to fuse question ($E_q$), condition ($E_c$) and image embeddings ($E_I$). It consists of four cross-attention stages. The first is \emph{Query to Condition attentions}. With $E_c$ as \emph{Key} and \emph{Value}, and all the three embedding as \emph{Query}, three multi-head attentions (MHA, Eq. \ref{equ:mha}) \cite{NIPS2017_attn} are employed. This stage simultaneously performs self-attention on $E_c$ and cross-attention on $E_q$ and $E_c$, and on $E_I$ and $E_c$. These interactions fully integrate $E_c$ with both other modalities and itself. The second is \emph{Query to Image attentions}, the same process as above is performed again with $E_I$ switch $E_c$ as \emph{Key} and \emph{Value}. Consequently, the three modalities further absorb desired information from the image side, effectively enabling IGTR to understand the known attributes and image content.
\begin{align}
\text{MHA}(Q, K, V) &= [\text{head}_1; \text{head}_2; \ldots; \text{head}_h] W^o \notag \\
\text{head}_i &= Attn_i(VW_i^v) \notag \\ 
Attn_i &= Softmax \left(\frac{(QW_i^q)(KW_i^k)^t}{\sqrt{D_h}} \right)
\label{equ:mha}  
\end{align}

In Eq. \ref{equ:mha}, the head number $h$ is set to 12, $W^o \in \mathbb{R}^{D \times D}$. The each head associates with three different weight matrices $W_i^q, W_i^k, W_i^v \in \mathbb{R}^{D \times D_h}$, where $D_h$ = $\frac{D}{h}$.

Notably, as shown in the upper right of Fig. \ref{fig:IGTR_overview}, three MHA modules are involved in each interaction, and their parameters are shared to ensure compact architecture and fast inference of IGTR. In this way, the two interactions above can be described by Eq. \ref{equ:equ2}:
\begin{align}
    \text{\emph{Query}} &= \left
    [ E_q;E_c;E_I \right] \in \mathbb{R}^{(L_q+L_c+\frac{HW}{32}) \times D} \notag \\
    \text{\emph{Query}}^1 &= LN(\text{MHA}(\text{\emph{Query}}, E_c, E_c) + \text{\emph{Query}}) \notag \\
    \left[E_q^1;E_c^1;E_I^1 \right] &= \text{\emph{Query}}^1 \notag \\
 \text{\emph{Query}}^2 &= LN(\text{MHA}(\text{\emph{Query}}^1, E_I^1, E_I^1) + \text{\emph{Query}}^1) \notag \\
 \left
    [E_q^2;E_c^2;E_I^2 \right] &= LN(FFN(\text{\emph{Query}}^2) + \text{\emph{Query}}^2) 
\label{equ:equ2}
\end{align}
\noindent where $LN$ means layer normalization.

In the following process (Eq. \ref{equ:equ3}), $E_q^2$ from Eq. \ref{equ:equ2} is employed as \emph{Query}, and question-to-condition and question-to-image cross-attentions are successively conducted. They correspond to the third and fourth cross-modality interactions, from which the question extracts finer features from both \emph{condition} and the image domain to obtain answer-related cross-modal embeddings $\hat{E}_q$.
\begin{align}
    E_q^{3} &= LN(\text{MHA}(E_q^2, E_c^2, E_c^2) + E_q^2) \notag \\
    E_q^{3} &= LN(FFN(E_q^{3}) + E_q^{3}) \notag \\
    E_q^{4} &= LN(\text{MHA}(E_q^{3}, E_I^2, E_I^2) + E_q^{3}) \notag \\
    \hat{E}_q &= LN(FFN(E_q^{4}) + E_q^{4}) 
\label{equ:equ3}
\end{align}

\noindent\textbf{Multi-task answer head.}
$\hat{E}_q$ represents the question embedding to be decoded and it can be split into $\hat{E}_{qc} \in \mathbb{R}^{N_c \times D}$, $\hat{E}_{qf} \in \mathbb{R}^{N_f \times D}$, $\hat{E}_{qp} \in \mathbb{R}^{N_p \times D}$, and $\hat{E}_{qs} \in \mathbb{R}^{N_s \times D}$ based on the four answer types. $N_c$, $N_f$, $N_p$, and $N_s$ are the number of answers corresponding to the four types and $L_q = N_c + N_f + N_p + N_s$. With those embeddings, a multi-task answer head is constructed to perform answer prediction. Specifically, the \emph{character} head, \emph{frequency} head, \emph{position} head, and \emph{status} head are responsible for questions whose answer is the character, frequency, position index, and status (\emph{Yes} or \emph{No}), respectively. The four heads are all linear classifiers, each associated with learnable parameter matrices: $W_{c} \in \mathbb{R}^{D \times C}$, $W_{f} \in \mathbb{R}^{D \times F}$, $W_{p} \in \mathbb{R}^{D \times 1}$, and $W_{s} \in \mathbb{R}^{D \times P},$ respectively. Consequently, the prediction answers $\hat{A}_{c}$, $\hat{A}_{f}$, $\hat{A}_{p}$, and $\hat{A}_{s}$ are obtained by applying the four heads to $\hat{E}_q$:
\begin{align}
    \hat{A}_{c} &= \sigma_1(\hat{E}_{qc} W_{c})\in \mathbb{R}^{N_c \times C}, \hat{A}_{f} = \sigma_1(\hat{E}_{qf} W_{f}) \in \mathbb{R}^{N_f \times F} \notag \\
    \hat{A}_{p} &= \sigma_1(\hat{E}_{qp} W_{p}) \in \mathbb{R}^{N_p \times P}, \hat{A}_{s} = \sigma_2(\hat{E}_{qs} W_{s}) \in \mathbb{R}^{N_s \times 1} \notag
% \label{equ:equ4}
\end{align}
\noindent where $\sigma_1$ and $\sigma_2$ are softmax and sigmoid functions.

\noindent\textbf{Answer loss.}
Prediction results of the four heads are compared with answer labels to calculate loss. Assuming $\hat{A}^n_{c}$ and ${A}^n_{c}$ are the prediction and answer label of the \emph{n}-th question in $Q_{c}$ and the rest defined similarity, the four losses are calculated as follows and then summed to get a total loss $\mathcal{L}$ according to the Eq. \ref{equ:equloss}:
\begin{align}
    \mathcal{L}_{c} &= \frac{1}{N_c} \sum_{n=0}^{N_c} \mathcal{L}_{ce}(\hat{A}_{c}^n, A_{c}^n, C) ,
    \mathcal{L}_{f} = \frac{1}{N_f} \sum_{n=0}^{N_f} \mathcal{L}_{ce}(\hat{A}_{f}^n, A_{f}^n, F) \notag \\
    \mathcal{L}_{s} &= \frac{1}{N_p} \sum_{n=0}^{N_p} \mathcal{L}_{ce}(\hat{A}_{p}^n, A_{p}^n, P), ~ 
    \mathcal{L}_{p} = \frac{1}{N_s} \sum_{n=0}^{N_s} \mathcal{L}_{bce}(\hat{A}_{s}^n, A_{s}^n) \notag \\
    \mathcal{L} &= \mathcal{L}_{c} + \mathcal{L}_{f} + \mathcal{L}_{s} + \mathcal{L}_{p}
\label{equ:equloss}
\end{align}
where cross-entropy ($\mathcal{L}_{ce}$) and binary cross-entropy ($\mathcal{L}_{bce}$) loss are defined as below.
\begin{align}
    \mathcal{L}_{ce}(\hat{Y}, Y, T)&=-\sum_{t=1}^{T}y_t\log(\hat{y}_t) \notag \\
\mathcal{L}_{bce}(\hat{Y}, Y)&=-y\log(\hat{y})-(1-y)\log(1-\hat{y})
\end{align}

\begin{figure}[t]
 \centering
\includegraphics[width=0.49\textwidth]{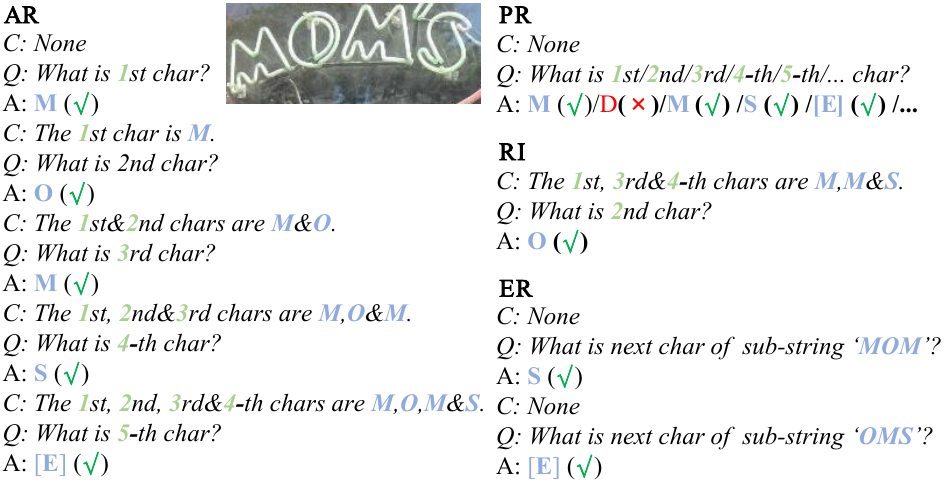} 
 \caption{The illustration of different recognition procedures by using different recognition instructions. [E] indicates the end symbol.}
 \label{fig:rec_process}
\end{figure}

\subsection{Text Recognition with instruction}

During model training, both the attribute prediction instructions in Tab. \ref{tab:instruction} and the recognition instructions in Tab. \ref{tab:rec_instruction} are employed. The attribute prediction instructions guide IGTR in learning fine-grained character-level and relational features, fostering a nuanced understanding of textual patterns. On the other hand, the recognition instructions, each corresponding to a different recognition pipeline, focus on equipping IGTR with the specific mechanisms for recognizing text. They together can enable the enhanced text image understanding capability.

When the model is sufficiently trained, during inference IGTR utilizes either the parallel recognition (PR), autoregressive recognition (AR), re-identification (RI), or extrapolated recognition (ER) instructions in Tab. \ref{tab:rec_instruction} to recognition text. Specifically, the first two instructions delineate the well-known PR and AR pipelines, respectively, standardized in an instruction-guided manner. Similar to existing methods \cite{yu2020srn,fang2021abinet,shi2019aster, Sheng2019nrtr}, both the two pipelines are limited to processing text whose length is up to 25 characters. The third instruction introduces RI. This instruction can guide the model to double-check the \emph{i}-th character, which is particularly useful when it is combined with PR or AR to recognize previously misidentified characters. In the fourth row, we propose ER, a novel pipeline somewhat analogous to AR. Starting from a given sub-string, each time it infers the previous and next characters of the sub-string, thus enabling a sub-string-based stepwisely extending recognition. It offers an elegant way to recognize text whose length is beyond the previous limit of 25 characters. In Fig. \ref{fig:rec_process}, we illustrate the recognition procedures by using the four types of instructions, which aid in elucidating their respective operational principles and implementation details.

\begin{table}[t]\footnotesize
\centering
\caption{Ablations on instruction variants, where accuracy (\%) of different recognition pipelines is given. The same below.}
\setlength{\tabcolsep}{6pt}{
\begin{tabular}{l|ccc}
\hline
Question (with all conditions)  & PR & AR & ER \\
\hline
(a):  recognition instructions                          & 80.57  & 81.27  & 80.79                 \\
(b): (a) + frequency attributes                           & 81.22   &  82.21  &  81.98              \\
(c): (b) + position attributes                           & 81.94 & 83.53  & 82.18                  \\
(d): (c) + sub-string attributes                          &  82.37 & 84.02   &  83.21                   \\
\hline
Condition (with all questions)  & PR & AR & ER                        \\
\hline
(e): w/o condition                                &  79.02    &  -  &  -                \\
(f): (e) + cond-1                        &  79.82  &  -  &  -                  \\
(g): (f) + cond-2                        &  80.90  &  -  &   -                 \\
(h): (g) + cond-3                        &  81.35  &  82.93  &   -                 \\
(i): (h) + cond-4                        &  82.23  &  83.97  & 83.25                   \\
\hline
K (with all conditions and questions) & PR & AR & ER                        \\
\hline
2                                     &  80.53    &  82.41  &  81.96                       \\
4                                     &  81.68    &  83.39  & 82.85                        \\
6                                     &  81.92    &  83.87  & 83.33                        \\
\hline
(j): ((d)+constraint, (i)+cond-5, K=8) & 82.51  &  84.86  & 83.78             \\
\hline
\end{tabular}}

\label{tab:instruct_ablation}
\end{table}

\section{Experiments}

\subsection{Datasets and Implementation Details}
\label{sec:Implementation}
We evaluate IGTR on both English and Chinese datasets. For English the employed training datasets include: (1) MJSynth (MJ) \cite{jaderberg14synthetic, Jader2015Reading} and SynthText (ST) \cite{Synthetic}, the two widely used synthetic scene text datasets. (2) Real-world Union14M-L training set, which contains over 3.2 million labeled real-world text images with both complexity and versatility \cite{jiang2023revisiting}. Note that Union14M-L training set overlaps with Union14M-L benchmark (test set) in 6,600 samples. These samples are filtered out from the training set to avoid data leakage. As for test protocols, the trained models are tested on: (1) six regular and irregular text benchmarks, i.e., ICDAR 2013 (IC13) \cite{icdar2013}, Street View Text (SVT) \cite{Wang2011SVT}, IIIT5K-Words (IIIT) \cite{IIIT5K}, ICDAR 2015 (IC15) \cite{icdar2015}, Street View Text-Perspective (SVTP) \cite{SVTP} and CUTE80 (CUTE) \cite{Risnumawan2014cute}. For IC13 and IC15, we use the versions with 857 and 1,811 images, respectively. We call the six benchmarks as Common benchmarks (abbreviated as Common). (2) Union14M-L benchmark \cite{jiang2023revisiting} (abbreviated as Union14M-L). It contains over 0.4 million test images. The benchmark is composed of seven challenging subsets including curve, multi-oriented, artistic, etc. For Chinese, we use Chinese text recognition (CTR) dataset \cite{chen2021benchmarking}, a benchmark containing four subsets: Scene, Web, Document, and Writing. We train the model on the whole training set and use the validation subset of Scene to determine the best model, which is then assessed on the four test subsets.

We use AdamW optimizer \cite{adamw} with a weight decay of 0.05 for training. For English models, all images are resized while maintaining their aspect ratio, with a maximum pixel count of $32 \times 128$ \cite{TPAMI2022ABINetPP, BautistaA22PARSeq, zheng2024cdistnet}. The learning rate (LR) is set to $5\times 10^{-4}$ and batchsize is set to 768. One cycle LR scheduler \cite{cosine} with 1.5 epochs linear warm-up is used in all the 20 epochs. 
The same as \cite{TPAMI2022ABINetPP, duijcai2022svtr, jiang2023revisiting}, data augmentation like rotation, perspective distortion, motion blur and gaussian noise, are randomly performed during training. The alphabet includes all case-insensitive alphanumerics. For Chinese models, all text instances are resized to $32 \times 256$ and data augmentation is not performed following \cite{yuICCV2023clipctr}. The LR is also set to $5\times 10^{-4}$ and batchsize is set to 512. One cycle LR scheduler with 3 epochs linear warm-up is used in all 100 epochs. Word accuracy is used as the evaluation metric. The size of the character set $C$ is set to 37 for English and 6625 for Chinese \cite{ppocrv3}. The maximum prediction length $L$ is set to 25 for both. All models are trained with mixed-precision on 2 Tesla A100 GPUs.

\begin{table}[t]\footnotesize
\centering
\caption{Ablations on CMFF components, where recognition accuracy (\%) and inference time (ms) are both given}.
\setlength{\tabcolsep}{2pt}{
\begin{tabular}{l|cccccc}
\hline
\multirow{2}{*}{Model} & \multicolumn{2}{c}{PR} & \multicolumn{2}{c}{AR} & \multicolumn{2}{c}{ER} \\
                       & Acc       & Time       & Acc       & Time       & Acc        & Time        \\
\hline
Full Model                    & 82.51 & 3.98 &  84.86 & 10.3 & 83.78 & 9.52                                   \\
(a): w/o Query to Condition    & 81.75  & 3.91 & 82.41 & 9.87 & 82.13        & 9.36                      \\
(b): w/o Query to Image          & 81.85 & 3.87 & 82.35        & 9.54 &  82.02            & 9.21                                     \\
(c): (a) + (b)                  &   80.85 & 3.77 & 81.23& 9.35 & 80.83     &   9.06                                         \\
(d): w/o Question to Condition &  81.96 & 3.93 & 82.22 & 9.86 & 82.01 & 9.39                               \\
(e): w/o Question to Image  &  81.34 & 3.90 & 83.21 & 9.53 & 82.88  &  9.24                          \\
(f): (d) + (e)  &  80.93     & 3.81 & 82.01 & 9.37 &   81.25 & 9.12
\\
\hline
\end{tabular}}
\label{tab:model_ablation}
\end{table}

\begin{figure}[t]
 \centering
\includegraphics[width=0.48\textwidth]{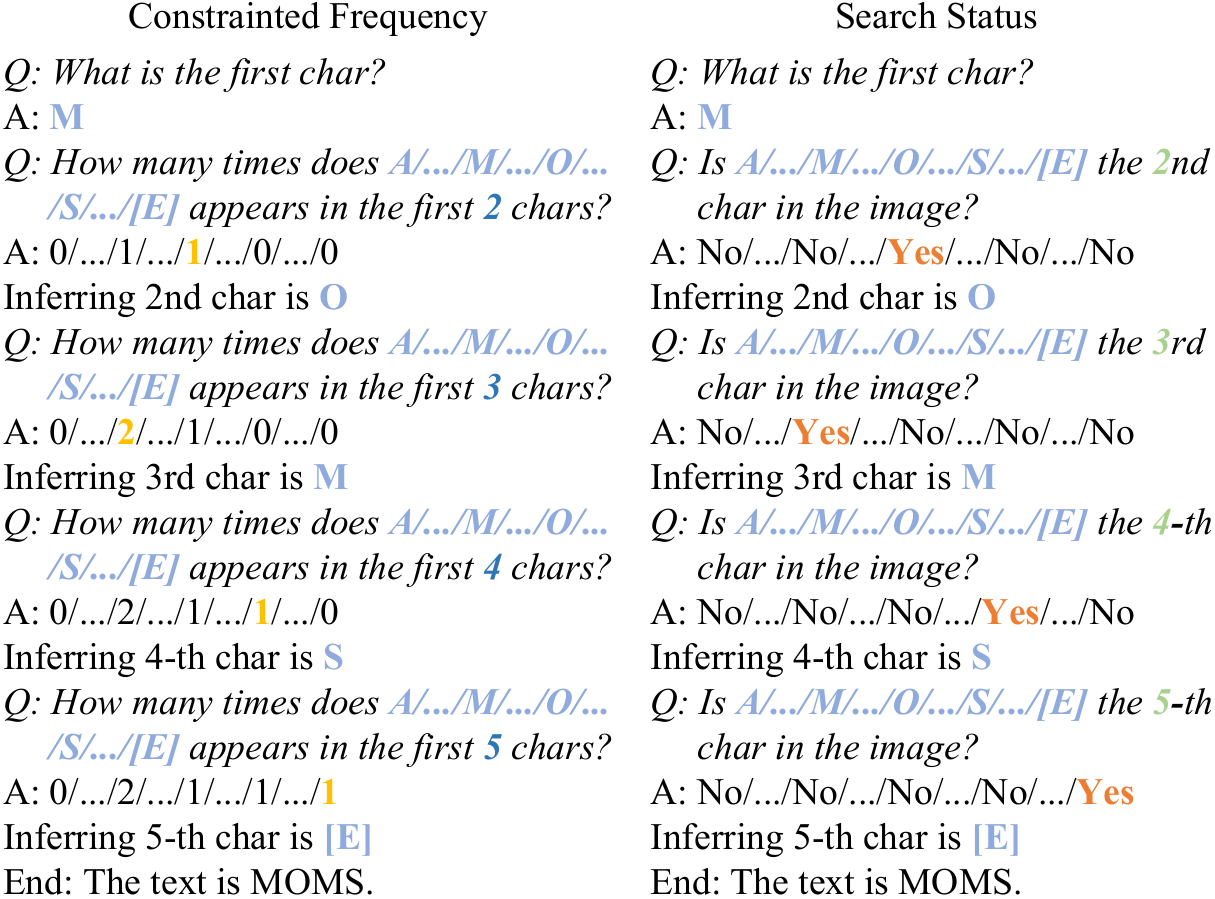} 
 \caption{Two illustrations of using different attribute prediction instructions described in Sec. 4.3 for text inference of image \emph{MOMS}. The $condition$ is set to None.}
 \label{fig:zeroshot}
\end{figure}

\subsection{Ablations} \label{section:3.3}

We conduct ablation studies to validate different instruction variants, CMFF components, and model scalability as follows. Note that all models are trained on Union14M-L training set and tested on its test set.

\noindent\textbf{Instruction variants.} 
We conduct experiments to validate the effectiveness of the triplet-form instructions. The results in Tab. \ref{tab:instruct_ablation} include IGTR-PR (PR), IGTR-AR (AR) and IGTR-ER (ER), where the suffix PR means using the PR instruction in Tab. \ref{tab:rec_instruction} for text inference and the others are defined similarly. In (a), only recognition instructions are utilized for IGTR training. In subsequent variants denoted by (b), (c), (d), and (j), we progressively add questions related to the referred attributes, where accuracy improvements are steadily obtained for all three models. The improvements reveal that the attributes all contribute positively, supporting that understanding diverse character attributes aids in recognition. Comparing the use of full instructions in (j) to (a), PR is improved by +1.94\%, AR by +3.59\%, and ER by +2.99\%, indicating that AR and ER benefit more from the diverse questions. The results align with the fact that compared to PR, AR and ER own more recognition clues. They employ an iterative pipeline and can utilize previously recognized characters as linguistic clues. On the other hand, we perform another set of ablations on \emph{condition}, starting from (e) with all questions but no condition ($cond$), and then progressively add a condition mode in (f)-(j), where the suffix is the condition index in Tab. \ref{tab:instruction}. IGTR enables AR and ER capabilities only when $cond$-3 and $cond$-4 are successively added, implying that enriching \emph{condition} not only enhances accuracy but also enables different recognition pipelines. Additionally, by increasing \emph{K}, the number of partitions, the accuracy steadily improved, again providing intuitive confirmation of the importance of improving instruction richness and diversity.

\noindent\textbf{CMFF components.} 
As shown in Fig. \ref{fig:IGTR_overview}, CMFF incorporates four cross-attention stages for inter-modal interaction. We devise variants by removing different stages, and the results are presented in Tab. \ref{tab:model_ablation}. Removing any stage all results in a decrease in accuracy. For example, when \emph{query-to-condition} and \emph{query-to-image} stages are both removed, notable accuracy declines of 1.66\% for PR, 3.63\% for AR, and 2.95\% for ER are observed. The results emphasize the necessity of incorporating both sufficient feature fusion stages and a single \emph{condition} embedding, which serves as \emph{Key} and \emph{Value} for the \emph{query-to-condition} stage. For AR and ER the \emph{condition} is preceding decoded characters while for PR is set to None. This explains why such removal has a more pronounced impact on AR and ER, while PR is relatively less affected. Meanwhile, when \emph{question-to-condition} and \emph{question-to-image} stages are both removed, the accuracy declines are 1.58\% for PR, 2.85\% for AR, and 2.53\% for ER, all preserving a consistent decreasing trend as above albeit with a smaller magnitude. This is because the two stages mainly focus on conducting finer feature interactions, and thus have relatively small affections on model performance. Furthermore, PR, AR, and ER report inference speeds of 3.98 ms, 10.3 ms and 9.52 ms, respectively. The speed differences align with their respective recognition pipelines. Note that they run faster than many previous models within their recognition pipelines. In addition, as shown in Tab. \ref{tab:model_ablation}, removing the cross-attention stages only triggers minor speed acceleration. The results, in turn, verify that CMFF is quite efficient.

\begin{table}[t]\footnotesize
\centering
\caption{Ablations on training data volume and model size.}
\setlength{\tabcolsep}{3pt}{
\begin{tabular}{c|ccc|c|ccc}
\hline
Epoch                      &  20 &  40 & 60 &  Model Size &  24M &  40M  \\
\hline
Vanilla-PR      &    76.14       &   76.89    &  77.05 & Vanilla-PR & 76.14  & 76.23 \\
IGTR-PR   &    82.51          &  84.06      & 85.29  & IGTR-PR & 82.51 & 83.60 \\
\hline
\end{tabular}}

\label{tab:scalability_ablation}
\end{table}

\begin{table}[t]\footnotesize
\centering
\caption{The recognition results of using different attribute prediction instructions.}
\setlength{\tabcolsep}{3.5pt}{
\begin{tabular}{l|cc}
\hline
Question                      &  Common &  Union14M-L \\
\hline
Constrainted frequency      &    94.43       &    80.22         \\
Search status   &     95.31         &    82.33      \\
\hline
\end{tabular}}

\label{tab:zeroshot_ablation}
\end{table}

\begin{table}[t]\footnotesize
\centering
\caption{Recognition results of rarely appearing characters on CTR \cite{chen2021benchmarking}.}
\setlength{\tabcolsep}{3.5pt}{
\begin{tabular}{l|cccc}
\hline
Model                      & $Rare_{1-10}$ & $Rare_{11-30}$ & $Rare_{31-50}$ & Avg \\
\hline
IGTR-PR      &    63.88       &    84.93 &  88.79 &  72.25         \\
IGTR-PR-TS   &     67.47         &    86.61  &  89.86 &  75.05     \\
\hline
\end{tabular}}

\label{tab:rare_char_tab}
\end{table}

\begin{table}[t]\footnotesize
\centering
\caption{Morphologically similar characters and mis-recognitions on CTR \cite{chen2021benchmarking} using IGTR-PR and IGTR-PR-TS. Take ``\ch{莱}-\ch{菜}" and 8/5 as the example. It means that without considering TS, ``\ch{莱}" has been recognized as ``\ch{菜}" for 8 times, and the number is reduced to 5 when TS is employed.}
\setlength{\tabcolsep}{2pt}{
\begin{tabular}{l|cccccc}
\hline
Char      & \ch{莱}      & x      & +        & \ch{太}     & \ch{干} & \ch{已}     \\
\hline
 & 8/5(\ch{菜}) & 8/4(×) & 22/14(\ch{十}) & 20/12(\ch{大}) & 13/11(\ch{千}) & 10/6(\ch{己}) \\
\hline
Char & \ch{入}     & \ch{土}    & \ch{域}    & \ch{曰}    & \ch{筒}    & z        \\
\hline
 & 12/8(\ch{人}) & 4/2(\ch{士}) & 6/3(\ch{城}) & 6/5(\ch{日}) & 6/4(\ch{简}) & 12/5(2)  \\
\hline
Char & \ch{金}     & \ch{全}    & \ch{内}    & \ch{于}    & \ch{天}    & \ch{凤}        \\
\hline
 & 26/15(\ch{全}) & 16/10(\ch{金}) & 15/8(\ch{肉}) & 13/8(\ch{子}) & 12/5(\ch{大}) & 12/9(\ch{风})  \\
\hline
Char & v     & \ch{东}    & i    & \ch{字}    & \ch{自}    & \ch{着}        \\
\hline
 & 22/15(y) & 7/4(\ch{乐}) & 106/86(l) & 9/4(\ch{宇}) & 8/5(\ch{白}) & 8/6(\ch{看})  \\
\hline
\end{tabular}}

\label{tab:siml_tab}
\end{table}

\begin{table*}[t]\footnotesize
\centering
\caption{Results on English benchmarks tested against existing models when trained on synthetic datasets. * represents that the result is evaluated on Union14M-L benchmarks using the model they released.}
\setlength{\tabcolsep}{2pt}{
\begin{tabular}{c|r|ccccccc|cccccccc|c}
\hline
\multicolumn{2}{c|}{\multirow{3}{*}{Method}} & \multicolumn{7}{c|}{Common Benchmarks}                                                                & \multicolumn{8}{c|}{Union14M-L Benchmark}                                                             & \multirow{3}{*}{\begin{tabular}[c]{@{}c@{}}Parameters\\  ($\times 10^6$)\end{tabular}} \\
\multicolumn{2}{c|}{}                        & IC13 & SVT  & IIIT & IC15 & SVTP & CUTE & Avg & Curve & \begin{tabular}[c]{@{}c@{}}Multi-\\ Oriented\end{tabular} & Artistic &\begin{tabular}[c]{@{}c@{}}Conte-\\ xtless\end{tabular}  & Salient & \begin{tabular}[c]{@{}c@{}}Multi-\\ Words\end{tabular} & General & Avg            &                           \\
\hline
\multirow{3}{*}{CTC}     & CRNN \cite{shi2017crnn}             & 91.1 & 81.6 & 82.9 & 69.4 & 70.0 & 65.5 & 76.75                                                                       & 7.5   & 0.9                                                       & 20.7     & 25.6        & 13.9    & 25.6                                                   & 32.0    & 18.03                                                                       & 8.3                                                                                                                 \\
& SVTR-B* \cite{duijcai2022svtr}          & 97.1 & 91.5 & 96.0 & 85.2 & 89.9 & 91.7 & 91.90                                                                       & 69.8  & \textbf{37.7}                                                      & 47.9     & 61.4        & 66.8    & 44.8                                                   & 61.0    & 55.63                                                                       & 24.6                                                                                                                \\
 & DCTC \cite{Zhang_2024_DCTC} & 97.4          & 93.7         & 96.9 & 87.3         & 88.5          & 92.3          & 92.68                      & -                 & -                                                      & -                 & -                 & -                 & -                                                   & -                 & -                                                & 40.8      \\
\hline
\multirow{17}{*}{AR}      & ASTER \cite{shi2019aster}           &   90.8   & 90.0 & 93.3 & 74.7 & 80.2 & 80.9 &  84.98                                                                           & 34.0  & 10.2                                                      & 27.7     & 33.0        & 48.2    & 27.6                                                   & 39.8    & 31.50                                                                       & 27.2                                                                                                                \\
& NRTR \cite{Sheng2019nrtr}            & 95.8 & 91.5 & 90.1 & 79.4 & 86.6 & 80.9 & 87.38                                                                       & 31.7  & 4.4                                                       & 36.6     & 37.3        & 30.6    & 54.9                                                   & 48.0    & 34.79                                                                       & 31.7                                                                                                                \\
& SAR \cite{li2019sar}             & 91.0 & 84.5 & 91.5 & 69.2 & 76.4 & 83.5 & 82.68                                                                       & 44.3  & 7.7                                                       & 42.6     & 44.2        & 44.0    & 51.2                                                   & 50.5    & 40.64                                                                       & 57.7                                                                                                                \\
& RoScanner \cite{yue2020robustscanner}   & 94.8 & 88.1 & 95.3 & 77.1 & 79.5 & 90.3 & 87.52                                                                       & 43.6  & 7.9                                                       & 41.2     & 42.6        & 44.9    & 46.9                                                   & 39.5    & 38.09                                                                       & 48.0                                                                                                                  \\
& PerSec \cite{Liu_2022_PerSec} & 97.2 & 94.6 & 96.3 & 84.4 & 89.5 & 90.2 & 92.03 &-&-&-&-&-&-&-& -    & - \\
& OpenCCD  \cite{Liu_2022_CVPR_OpenCCD}   & 92.2          & 85.9          & 91.9 & -          & -          & 83.9          & -                                                                   &-&-&-&-&-&-&-& -    & -                                 \\
& SGBANet  \cite{sgbanet}  & 95.1          & 89.1          & 95.4 & -          & 83.1          & 88.2          & -                                                                &-&-&-&-&-&-&-&  -      & -                        \\
& PARSeq* \cite{BautistaA22PARSeq}          & 97.0 & 93.6 & 97.0 & 86.5 & 88.9 & 92.2 & 92.53                                                                       & 63.9  & 16.7                                                      & 52.5     & 54.3        & 68.2    & 55.9                                                   & 56.9    & 52.62                                                                       & 23.8         \\
& CornerTrans*  \cite{xie2022toward}   & 97.8          & 94.6          & 95.9 & 86.5          & 91.5          & 92.0          & 93.05                                                                  &62.9&18.6&56.1&58.5&68.6&59.7&61.0&   55.07   & 86.0                       \\

& LevOCR*  \cite{levocr}  & 96.7          & 94.4          & 96.6 & 86.5          & 88.8          & 90.6          & 92.27                                                                    &52.8&10.7&44.8&51.9&61.3&54.0& 58.1&   47.66 & 109                          \\
& DiG \cite{YangLLWZLTB22_DiG} & 96.9 & 94.6 & 96.7 & 87.1 & 91.0 &91.3 & 92.93 &-&-&-&-&-&-&-&  -      & -\\
& SIGA* \cite{Guan_2023_CVPR_SIGA}   & 97.8                 & 95.1                                                      & 96.6                 & 86.6                 & 90.5                 & 93.1                                     & 93.28 &59.9&22.3&49.0&50.8&66.4&58.4&56.2& 51.85 & 113   \\ 
& CCD* \cite{Guan_2023_ICCV_CCD}   & 97.0                 & 94.4                                                      & 97.2                 & 87.6                 & 91.8                 & 93.3                                     & 93.55 &66.6&24.2&63.9&64.8&74.8&62.4&64.0& 60.10   & 52.0    \\ 
& LISTER* \cite{iccv2023lister} & 97.9 & 93.8 & 96.9 & 87.5 & 89.6 & 90.6 & 92.72 & 56.5 & 17.2 & 52.8 & 63.5 & 63.2 & 59.6 & 65.4 & 54.05 & 49.9 \\
& CDisNet* \cite{zheng2024cdistnet}         & 97.4 & 93.5 & 96.4 & 86.0 & 88.7 & 93.4 & 92.57                                                                       & 69.3  & 24.4                                                      & 49.8     & 55.6        & 72.8    & 64.3                                                   & 58.5    & 56.38                                                                       & 65.5                                                                                                                \\
   & CAM* \cite{yang2024class_cam} & 97.2          & \textbf{96.1}         & 97.4 & 87.8         & 90.6          & 92.4          & 93.58                      & 63.1                 & 19.4                                                      & 55.4                 & 58.5                 & 72.7                 & 51.4                                                   & 57.4                 & 53.99                                               & 135      \\
 & OTE \cite{Xu_2024_CVPR_OTE} & 97.4          & 95.5         & 96.4 & 87.2         & 89.6          & 92.4          & 93.08                      & -                 & -                                                      & -                 & -                 & -                 & -                                                   & -                 & -                                               & 25.2      \\
\hline
\multirow{9}{*}{PR}      & SRN \cite{yu2020srn}             & 95.5 & 91.5 & 94.8 & 82.7 & 85.1 & 87.8 & 89.57                                                                       & 63.4  & 25.3                                                      & 34.1     & 28.7        & 56.5    & 26.7                                                   & 46.3    & 40.14                                                                       & 54.7                                                                                                                \\
& VisionLAN \cite{Wang_2021_visionlan}       & 95.7 & 91.7 & 95.8 & 83.7 & 86.0 & 88.5 & 90.23                                                                       & 57.7  & 14.2                                                      & 47.8     & 48.0        & 64.0    & 47.9                                                   & 52.1    & 47.39                                                                       & 32.8                                                                                                                \\
& ABINet \cite{TPAMI2022ABINetPP}          & 97.4 & 93.5 & 96.2 & 86.0 & 89.3 & 89.2 & 91.93                                                                       & 59.5  & 12.7                                                      & 43.3     & 38.3        & 62.0    & 50.8                                                   & 55.6    & 46.03                                                                       & 36.7                                                                                                                \\
& GTR \cite{HeC0LHWD22GTR}             & 96.8 & 94.1 & 95.8 & 84.6 & 87.9 & 92.3 & 91.92                                                                       & 62.3  & 13.9                                                      & 50.0     & 45.1        & 67.1    & 53.4                                                   & 58.5    & 50.07                                                                       & 42.1                                                                                                                \\
& MGP-STR*  \cite{mgpstr}     & 97.3          & 94.7         & 96.4 & 87.2         & 91.0          & 90.3          & 92.82                                                                     &55.2 &14.0& 52.8&48.5&65.2&48.8& 59.1& 49.09  & 148       \\
& MATRN  \cite{MATRN}       & 97.9          & 95.0         & 96.6 & 86.6         & 90.6          & 93.5          & 93.37                      & 63.1                 & 13.4                                                      & 43.8                 & 41.9                 & 66.4                 & 53.2                                                   & 57.0                 & 48.40                                                & 44.2      \\

& LPV-B* \cite{ijcai2023LPV}           & 97.6 & 94.6 & 97.3 & 87.5 & 90.9 & 94.8 & 93.78                                                                       & 68.3  & 21.0                                                      & 59.6     & 65.1        & 76.2    & 63.6                                                   & 62.0    & 59.40                                                                       & 35.1                                                                                                                \\
&CPPD \cite{du2023cppd}& 98.4 & 95.8 & 97.5 & 88.3 & 91.6 & 92.7 & 94.06   & 68.7    & 18.8     & 56.5    & 60.9    & 72.4  & 59.1   & 65.5   & 57.40    & 46.2 \\
 & BUSNet \cite{Wei_2024_busnet} & 98.3          & 95.5         & 96.2 & 87.2         & 91.8         & 91.3          & 93.38                      & -                 & -                                                      & -                 & -                 & -                 & -                                                   & -                 & -                                               & 56.8      \\

\hline
\multirow{4}{*}{Ours}    & IGTR-PR        & 97.6     &  95.2    &   97.6   &  88.4    &    91.6  &  \textbf{95.5}    & 94.30                                                                        &  76.9     &  30.6                                                         &      59.1    &    63.3         &     77.8    &                                                  62.5      &     66.7    & 62.40                                                                        & 24.1                                                                                                                  \\
& IGTR-AR        & \textbf{98.6}     &  95.7    &   \textbf{98.2}   &    88.4  &   \textbf{92.4}   &  \textbf{95.5}    &  \textbf{94.78}                                                                           &  \textbf{78.4}     &    31.9                                                       &      \textbf{61.3}    &   \textbf{66.5}          &   \textbf{80.2}      &    \textbf{69.3}                                                    &  \textbf{67.9}       &    \textbf{65.07}                                                                         &   24.1
                         \\
&IGTR-PR-RI        & 97.7     &  95.5    &   97.7   & \textbf{88.5}    &    91.6  &  \textbf{95.5}    & 94.43                                                                        &  77.3     &  31.0                                                         &      59.6    &    64.3         &     78.4    &                                                  65.9      &     67.2    & 63.40                                                                        & 24.1                                                                                                                  \\
&IGTR-ER       &  97.3    &  94.9    &   97.2    &   88.3     & 91.7     &    95.1  & 94.09                                                                       &  78.2     &     32.0      & 60.6      &   59.1          &     78.2    &  57.8                                                      &  67.1       & 61.84                                                                        & 24.1                                                                  \\
\hline
\end{tabular}}
\label{tab:sota_syn}
\end{table*}

\begin{table*}[t]\footnotesize
\centering
\caption{Results on English benchmarks tested against existing models when trained on real-world Union14M-L training set. $^\dag$ denotes that the result is obtained by training the model on Union14M-L using the code they released.}
\setlength{\tabcolsep}{2pt}{
\begin{tabular}{c|r|ccccccc|cccccccc|c}
\hline
\multicolumn{2}{c|}{\multirow{3}{*}{Method}} & \multicolumn{7}{c|}{Common Benchmarks}                                                                & \multicolumn{8}{c|}{Union14M-L Benchmark}                                                             & \multirow{3}{*}{\begin{tabular}[c]{@{}c@{}}Parameters\\  ($\times 10^6$)\end{tabular}} \\
\multicolumn{2}{c|}{}                        & IC13 & SVT  & IIIT & IC15 & SVTP & CUTE & Avg & Curve & \begin{tabular}[c]{@{}c@{}}Multi-\\ Oriented\end{tabular} & Artistic & \begin{tabular}[c]{@{}c@{}}Conte-\\ xtless\end{tabular} & Salient & \begin{tabular}[c]{@{}c@{}}Multi-\\ Words\end{tabular} & General & Avg            &                           \\
\hline
\multirow{2}{*}{CTC}     & CRNN \cite{shi2017crnn}             & 91.8 & 83.8 & 90.8 & 71.8 & 70.4 & 80.9 & 81.58                                                                       & 19.4  & 4.5                                                       & 34.2     & 44.0        & 16.7    & 35.7                                                   & 60.4    & 30.70                                                                       & 8.3                                                                                                                 \\
& SVTR-B$^\dag$ \cite{duijcai2022svtr}          & 97.5 & 96.4 & 97.8 & 89.3 & 91.0 & 96.2 & 94.72                                                                       & 85.4  & 87.4                                                      & 68.9     & 79.5        & 84.3    & 79.1                                                   & 81.8    & 80.91                                                                       & 24.6                                                                                                                \\
\hline
\multirow{7}{*}{AR}      & ASTER \cite{shi2019aster}            & 92.6 & 88.9 & 94.3 & 77.7 & 80.5 & 86.5 & 86.75                                                                       & 38.4  & 13.0                                                      & 41.8     & 52.9        & 31.9    & 49.8                                                   & 66.7    & 42.07                                                                       & 27.2                                                                                                                \\
& NRTR \cite{Sheng2019nrtr}             & 96.9 & 94.0 & 96.2 & 80.9 & 84.8 & 92.0 & 90.80                                                                       & 49.3  & 40.6                                                      & 54.3     & 69.6        & 42.9    & 75.5                                                   & 75.2    & 58.20                                                                       & 31.7                                                                                                                \\
& SAR \cite{li2019sar}              & 96.0 & 92.4 & 96.6 & 82.0 & 85.7 & 92.7 & 90.90                                                                       & 68.9  & 56.9                                                      & 60.6     & 73.3        & 60.1    & 74.6                                                   & 76.0    & 67.20                                                                       & 57.7                                                                                                                \\
& RoScanner \cite{yue2020robustscanner}    & 95.7 & 92.4  & 96.8 & 86.4 & 83.9 & 93.8 & 91.50                                                                       & 66.2  & 54.2                                                      & 61.4     & 72.7        & 60.1    & 74.2                                                   & 75.7    & 66.36                                                                       & 48.0                                                                                                                  \\
& MAERec \cite{jiang2023revisiting}        & 97.6 & 96.8 & 98.0  & 87.1 & 93.2 & 97.9 & 95.10                                                                       & 81.4  & 71.4                                                      & 72.0     & 82.0        & 78.5    & 82.4                                                   & 82.5    & 78.60                                                                       & 35.7                                                                                                                \\
&   LISTER$^\dag$ \cite{iccv2023lister}   & 97.4 & 98.1 & 98.2 & 89.2 & 93.5 & 95.5 & 95.33          & 71.6 & 55.9 & 68.9 & 76.4 & 68.1 & 80.2 & 80.9 & 71.72  & 49.9 \\
 & OTE \cite{Xu_2024_CVPR_OTE}   & 98.0 & 98.0 & 98.1 & 89.1 & 95.5 & 97.6 & 96.10          & 83.1 & 82.8 & 73.5 & 73.7 & 79.7 & 70.3 & 82.2 & 77.90     & 25.2 \\
\hline
\multirow{5}{*}{PR}      & SRN \cite{yu2020srn}              & 94.7 & 89.5 & 95.5 & 79.1 & 83.9 & 91.3 & 89.00                                                                       & 49.7  & 20.0                                                      & 50.7     & 61.0        & 43.9    & 51.5                                                   & 62.7    & 48.50                                                                       & 54.7                                                                                                                \\
& VisionLAN \cite{Wang_2021_visionlan}    & 95.1 & 91.3 & 96.3  & 83.6 & 85.4 & 92.4 & 90.68                                                                       & 70.7  & 57.2                                                      & 56.7     & 63.8        & 67.6    & 47.3                                                   & 74.2    & 62.50                                                                       &  32.8                                                                                                                \\
& ABINet \cite{TPAMI2022ABINetPP}       & 97.2 & 95.7 & 97.2 & 87.6 & 92.1 & 94.4 & 94.03                                                                       & 75.0  & 61.5                                                      & 65.3     & 71.1        & 72.9    & 59.1                                                   & 79.4    & 69.19                                                                       & 36.7                                                                                                               \\       
& MATRN \cite{MATRN}       & 97.9 & 96.9 & 98.2 & 88.2 & 94.1 & 97.9 & 95.50                                                                       & 80.5  & 64.7                                                      & 71.1     & 74.8        & 79.4    & 67.6                                                   & 77.9    & 74.60     & 44.2                                      \\ 
& LPV-B$^\dag$ \cite{ijcai2023LPV}	& 97.4 	& 97.4 	& 98.9 	& 89.8 	& 93.0 	& 97.2 	& 95.62 	& 82.4 	& 64.6 	& 74.1 	& 81.0 	& 78.8 	& 81.1 	& 82.8 	& 77.83 & 30.5\\
\hline
\multirow{6}{*}{Ours}    & IGTR-PR        &  97.7    &   97.7   &  98.3    &  89.8    &  93.7    &  97.9    & 95.86                                                                       &   88.1   & 89.9    &     74.2    &     80.3      &   82.8      &   79.2    &   83.0      & 82.51                                                                        & 24.1                                                                                                                  \\
& IGTR-AR       &  98.1    &  \textbf{98.4}    &   98.7    &   90.5     & 94.9     &    98.3  & 96.48                                                                       &  90.4     &     91.2      & 77.0      &   82.4          &     84.7    &  84.0                                                      &  84.4       & 84.86                                                                        & 24.1                                                                                                                  \\
&IGTR-PR-RI & 97.8    &   97.8   &  98.3    &  89.7    &  93.8    &  97.9    & 95.91                                                                       &   88.6   & 90.0    &     74.1    &     80.6      &   83.2      &   79.9    &   83.5      & 82.86                                                                        & 24.1 \\
&IGTR-ER       &  98.1    &   97.8    &   98.3    &   90.5     & 94.0     &    97.2  & 95.99                                                                       &  89.4     &     92.3      & 76.2      &  78.9          &     84.7    &  80.9                                                      &  84.0       & 83.78                                                                        & 24.1                                                                  \\

& IGTR-PR-PT    &  98.6    &   98.0   &   99.1   &    91.7  &  \textbf{96.8}    & \textbf{99.0}     & 97.20                                                                        &   92.4    &       92.1                                                    &     80.7     &      \textbf{83.6}       &   87.7      &     86.9                                                   &   85.0      & 86.92                                                                        & 24.1                                                                                                                  \\

& IGTR-AR-PT    & \textbf{98.8}     &   98.3   &   \textbf{99.2}   &  \textbf{92.0}    &  \textbf{96.8}    &   \textbf{99.0}   & \textbf{97.34}                                                                        &  \textbf{93.0}     &   \textbf{92.9}                                                        &        \textbf{81.3}   &     83.4       &   \textbf{88.6}      &    \textbf{88.7}                                                    &    \textbf{85.6}     & \textbf{87.65}                                                                        & 24.1               \\
\hline
\end{tabular}}

\label{tab:sota_u14m}
\end{table*}

\noindent\textbf{Scalability of IGTR.} 
We conduct ablations on training data volume and model size. Specifically, we select IGTR-PR and vanilla-PR, where the latter is trained solely using the PR instruction in Tab. \ref{tab:rec_instruction}. The results are presented in the left half of Tab. \ref{tab:scalability_ablation}, where different training epochs are considered. We observe a notable accuracy gain of 2.78\% when increasing epochs from 20 to 60 for IGTR-PR. This improvement is accompanied by the expansion of training data volume, as IGTR-PR samples attribute prediction instructions individually at each epoch. The trained model benefits from richer instructions. In contrast, the improvement of vanilla-PR is 0.91\%, solely from more training epochs that generate a slightly better model optimization. The significant improvement gap (2.78\% v.s. 0.91\%) implies that our instruction-guided training is more flexible and can produce more powerful STR models when necessary. Furthermore, the right half of Tab. \ref{tab:scalability_ablation} shows that the accuracy increases with model size. The 24M and 40M models correspond to using SVTR-B and SVTR-L \cite{duijcai2022svtr} as the image encoder, with the remaining architecture keeping the same. We observe a similar large improvement gap (1.09\% v.s. 0.09\%) between the two methods, suggesting that diverse attribute prediction instructions have a greater potential to enhance the capability of the visual backbone. However, to fairly compare with other models, we take SVTR-B as the backbone and train IGTR models with fixed 20 epochs in subsequent evaluations.

\subsection{Recognition by Attribute Prediction Instructions}

It is observed that the attribute prediction instructions, when intelligently combined, can also infer the whole text. We refer to this pipeline as attribute prediction-based recognition. To validate this pipeline, we first train the model on attribute prediction instructions, then employ two questions to enable text inference. They are \emph{constrainted frequency} and \emph{search status}. The quantitative results are given in Tab. \ref{tab:zeroshot_ablation}, and Fig. \ref{fig:zeroshot} illustrates the recognition procedures. The former question, when combined with another already learned \emph{edge char} question that tells the first character, enables a step-by-step inference of subsequent characters, thus recognizing all characters. Specifically, we first set the second variable in question \emph{constrainted frequency} to \emph{2} and enumerate all the characters, from which the second character can be determined. Then, the third character can be deduced as similar by modifying the second variable in the question to \emph{3}, and so on. However, this inference chain is relatively complex and the model only achieves moderate success on challenging datasets (80.22\% on Union14M-L). The latter question, by traversing all characters across all positions, presents a more straightforward and less chain-dependent way to determine the text. It reaches an accuracy of 82.33\% on Union14M-L. The results demonstrate that by training solely on attribute prediction instructions, without the inclusion of text recognition instructions, IGTR can inherently develop certain character-level recognition capabilities. The text is then recognized when questions in attribute prediction instructions are appropriately utilized.

\begin{CJK}{UTF8}{gbsn}

\subsection{Rare and Similar Character Recognition}

Recognizing rarely appearing and morphologically similar characters are two typical challenges in STR, especially for Chinese which has a large character vocabulary. Owning to the flexibility of instruction-guided learning, we devise an elegant scheme to alleviate the two challenges by adding certain rules to instruction sampling. The first part is for rare characters. Specifically, we first inspect the CTR dataset \cite{chen2021benchmarking}, characters with less than 50 occurrences in the validation set are treated as rare characters. When a text instance contains a rare character $c_r$, attributes related to $c_r$ are used to generate question-answer pairs, while other attributes are randomly divided into \emph{condition} and \emph{question-answer} pairs as before. In addition, for rare characters that are not in the input text, five of them are randomly selected and are used to generate questions of \emph{search status}. For example, \emph{Is $c_i$ the j-th character in the image?}, where $c_i$ is one selected rare character. By adjusting the sampling rules as above, attributes related to rare characters receive more attention, and thus learned better. 

Then we explain how the rule is adjusted for morphologically similar characters. For a given character, we randomly select three characters, which the given character has the most frequently been misclassified to, as its morphologically similar characters, or all characters are selected if the misclassified characters are less than three. Then, given a text instance, specific \emph{search status} questions are sampled for appeared characters in the form of these similar character pairs. For example, for a similar character pair ``大-太", where the first character is the appeared character and the second is its morphologically similar character. The answer is \emph{Yes} when the question is \emph{Is the \(i\)-th character ``大"?}, and \emph{No} when the question is \emph{Is the \(i\)-th character ``太"?}. By reinforcing these facts during model training, morphologically similar characters can be better distinguished.

We term the two adjustments above as Targeted Strengthening (TS) and devise a model termed IGTR-PR-TS to incorporate them. Results on the CTR dataset \cite{chen2021benchmarking} are presented in Tab. \ref{tab:rare_char_tab} and Tab. \ref{tab:siml_tab}, respectively. In Tab. \ref{tab:rare_char_tab}, rare characters are grouped into three classes according to their occurrences. Compared to the raw IGTR-PR, the accuracy improvements are 3.59\%, 1.68\%, and 1.07\% for characters appearing 1 to 10, 11 to 30, and 31 to 50 times, respectively. Larger gains are observed for severely rare characters. Meanwhile in Tab. \ref{tab:siml_tab}, each character pair gives two values, e.g., 8/5 for the ``莱-菜" pair. It means that IGTR-PR classifies ``莱" as ``菜" 8 times, and the times is reduced from 8 to 5 when IGTR-PR-TS is employed. As can be seen, recognition errors occurred between morphologically similar characters are largely reduced. Both experiments convincingly verify the flexibility of IGTR and its potential to address typical STR challenges.

\end{CJK}

\subsection{Comparison with State-of-the-Art}

\begin{table}[t]\footnotesize
\centering
\caption{Results on the Occluded Scene Text dataset \cite{Wang_2021_visionlan}. All the models are trained on Union14M-L training set except for those suffixed with PT.}
\label{tab:ost}

\setlength{\tabcolsep}{5pt}{
\begin{tabular}{c|cccccccc}
\hline
& Vanilla-PR & IGTR-PR &IGTR-AR & IGTR-PR-PT   \\
\hline
WOST & 82.5 & 83.2 & 85.1 & 89.7   \\ 
HOST & 64.1 & 65.3 & 67.6 & 82.1 \\
\hline
Avg  & 73.30 & 74.25 & 76.32 & 85.90 \\
\hline
\hline
&IGTR-AR-PT & SVTR-B \cite{duijcai2022svtr} & LPV-B \cite{ijcai2023LPV} & MAERec \cite{jiang2023revisiting} \\
\hline
WOST& 90.2 & 80.6 & 80.7 & 79.1 \\
HOST& 82.7 & 65.0 & 65.9 & 65.2  \\
\hline
Avg& 86.45 & 72.8 & 73.30 & 72.15 \\
\hline
\end{tabular}}
\end{table}

\begin{table}[t]\footnotesize
\centering
\caption{Evaluations on pre-training methods, where the attribute prediction-based pre-training is substituted by three existing STR pre-training schemes. Com and U14M denote Common and Union14M-L benchmarks, respectively.}
\label{tab:pretrained}

\begin{tabular}{c|cc|cc|cc}
\hline
\multirow{2}{*}{\begin{tabular}[c]{@{}c@{}}Pretraining\\ Methods\end{tabular}} & \multicolumn{2}{c|}{SeqCLR \cite{AberdamLTASMMP21_SeqCLR}} & \multicolumn{2}{c|}{MAERec \cite{jiang2023revisiting}} & \multicolumn{2}{c}{DiG \cite{YangLLWZLTB22_DiG}} \\
& Com          & U14M        & Com        & U14M       & Com          & U14M        \\
\hline
IGTR-PR                                                                     & 96.36        & 83.73    & 96.60        & 84.34   & 96.51      & 84.91             \\
IGTR-AR                                                                     & 96.71        & 85.36             & 96.85        & 86.30    & 96.90      & 86.74  \\
\hline
\end{tabular}
\end{table}

\begin{table}[t]\footnotesize
\centering
\caption{Results on CTR dataset tested against existing models.}
\setlength{\tabcolsep}{2.5pt}{
\begin{tabular}{r|cccc|c|c}
\hline
Method         & Scene & Web & \begin{tabular}[c]{@{}c@{}}Docu-\\  ment\end{tabular} & \begin{tabular}[c]{@{}c@{}}Hand-\\  writing\end{tabular} & 
Avg & \begin{tabular}[c]{@{}c@{}}Params\\  ($\times 10^6$)\end{tabular}                              \\
\hline
CRNN \cite{shi2017crnn}          & 53.4  & 57.0 & 96.6 & 50.8 & 64.45   & 12.4                          \\
ASTER \cite{shi2019aster}         & 61.3  & 51.7 & 96.2 & 37.0 &  61.55  & 27.2                            \\
MORAN \cite{pr2019MORAN}         & 54.6  & 31.5 & 86.1 & 16.2  &  47.10  & 28.5                            \\
SAR \cite{li2019sar} & 59.7     & 58.0 & 95.7 & 36.5  & 62.48  & 27.8                           \\
SEED \cite{cvpr2020seed} & 44.7     & 28.1 & 91.4 & 21.0  &   46.30 & 36.1                           \\
MASTER \cite{pr2021MASTER}         & 62.8 & 52.1 & 84.4 & 26.9 & 56.55 & 62.8                                 \\
ABINet \cite{TPAMI2022ABINetPP}         & 66.6 & 63.2 & 98.2 & 53.1 & 70.28  & 53.1                              \\
TransOCR \cite{cvpr2021TransOCR}        & 71.3 & 64.8 & 97.1 & 53.0 & 71.55  & 83.9                               \\           
SVTR-B$^\dag$ \cite{duijcai2022svtr}         & 71.7 & 73.8 & 98.2 & 52.2 & 73.98  & 26.3 \\
CCR-CLIP \cite{yuICCV2023clipctr}         & 71.3 & 69.2 & 98.3 & 60.3 & 74.78  & 62.0   \\
DCTC \cite{Zhang_2024_DCTC}     & 73.9          & 68.5          & 99.4          & 51.0          & 73.20          & 40.8     \\
CAM \cite{yang2024class_cam}     & 76.0          & 69.3          & 98.1          & 59.2          & 76.80          & 135     \\
\hline
IGTR-PR      & 73.1 & 74.8 & 98.6 &  52.5 & 74.75  & 29.2 \\
IGTR-AR    & 75.1 & 76.4 & 98.7 &  55.3 & 76.37  & 29.2  \\
IGTR-PR-TS     & 73.5 & 75.9 & 98.7 &  54.5 & 75.65  & 29.2  \\
IGTR-AR-TS     & 75.6 & 77.0 & 98.8 &  57.3 & 77.17  & 29.2 \\
IGTR-PR-TS-Aug     & 79.5 & 80.0 & 99.4 &  58.9 & 79.45  & 29.2  \\
IGTR-AR-TS-Aug     & \textbf{82.0} & \textbf{81.7} & \textbf{99.5} &  \textbf{63.8} & \textbf{81.74}  & 29.2 \\
\hline
\end{tabular}}

\label{tab:chinese}
\end{table}

We compare IGTR with previous STR models on both Common and Union14M-L benchmarks. The results trained based on synthetic and Union14M-L training set are presented in Tab.~\ref{tab:sota_syn} and Tab.~\ref{tab:sota_u14m}, respectively. Meanwhile, we also train IGTR in another manner: first pre-trained with attribute prediction instructions on the synthetic datasets, and then fine-tuned with text recognition instructions on Union14M-L training set. The obtained models are marked with a suffix PT, e.g., IGTR-PR-PT and IGTR-AR-PT.

We first inspect the results trained on synthetic datasets. As shown in Tab.~\ref{tab:sota_syn}, in \emph{Ours} we give results of four IGTR models corresponding to the four text recognition pipelines. Note that IGTR-PR-RI means RI is further incorporated into the IGTR-PR pipeline. IGTR-AR ranks the top among 11 of the 13 evaluated subsets from both Common and Union14M-L benchmarks. It surpasses IGTR-PR by 0.48\% on Common and 2.67\% on Union14M-L in terms of accuracy, respectively, demonstrating the effectiveness of seeing already decoded characters during recognition, especially on more challenging Union14M-L. For IGTR-PR-RI v.s. IGTR-PR, 1\% improvement is observed on Union14M-L, showing that the re-identification can correct misrecognition to some extent. Moreover, the four IGTR models perform almost better than all the compared existing models in Tab.~\ref{tab:sota_syn}. For example, IGTR-AR outperforms LPV-B \cite{ijcai2023LPV}, one of the most competitive previous models, by 1.0\% and 5.67\% on Common and Union14M-L, respectively. Those results convincingly indicate the superiority of our comprehension-first and recognition-next paradigm.

We then examine the models trained on real-world datasets. The results on Union14M-L are depicted in Tab.~\ref{tab:sota_u14m}. All models show a significant improvement in accuracy when trained on Union14M-L training set, underscoring the importance of real-world training data. Similarly, IGTR models excel all existing models in Tab.~\ref{tab:sota_u14m} except for OTE \cite{Xu_2024_CVPR_OTE} in a few subsets within Common. IGTR-PR-PT and IGTR-AR-PT, the two models mixed synthetic data-based pre-training and real-world data-based fine-tuning, perform considerably well. IGTR-AR-PT achieves the highest rank in 11 out of the 13 evaluated subsets. The results showcase that the first pre-training and then fine-tuning scheme is a superior training paradigm. This can be explained as more training data is involved, and meanwhile, this paradigm follows an easy-to-hard learning procedure, i.e., first pre-trains on less challenging synthetic data and then fine-tunes on more difficult real-world data. For IGTR-AR-PT, the accuracy gaps to SVTR-B, one of the best previous models in Tab.~\ref{tab:sota_u14m}, are prominent 2.62\% on Common and 6.74\% on Union14M-L, respectively. The results again demonstrate the superiority of our instruction-guided learning. When looking back to the attribute prediction-based recognition results in Tab. \ref{tab:zeroshot_ablation}, we observe that the results obtained based on the \emph{search status} question already surpass existing top-performed models on Union14M-L. This implies the effectiveness of the devised character attribute prediction instructions, which already endow the model with comprehensive text understanding. Note that all IGTR models only differ in training details. These models are all with a model size of 24.1M, which is highly competitive compared to existing models.

We further analyze the performance of IGTR in \textit{Contextless}, a subset in Union14M-L containing text that has no semantic meaning and is not in the dictionary. The results are listed in Tab. \ref{tab:sota_u14m} and from which two observations can be inferred. First, IGTR models still get quite impressive accuracy. The accuracy of IGTR-PR-PT reaches 83.6\% while IGTR-AR-PT achieves 83.4\%. Both results surpass previous leading methods such as LPV-B \cite{ijcai2023LPV} and MAERec \cite{jiang2023revisiting}. Since contextless text provides limited linguistic prior, the accuracy gains are mainly because the rich and diverse instructions compel the model to learn character attributes from different aspects during IGTR training. Thus fine-grained visual content has been well understood and robust visual recognition capability is established. Second, IGTR-PR-PT performs slightly better than IGTR-AR-PT, which is opposed to the comparison on the other six subsets in Union14M-L. This implies that the availability of previously decoded characters in the AR pipeline does not contribute to the recognition, which also aligns with the text's contextless nature. Both observations indicate IGTR's superior text recognition capability, and IGTR-PR-PT and IGTR-RR-PT comprehensively understand character attributes after the pre-training.

To further assess the robustness of IGTR models, we conduct experiments on Occluded Scene Text (OST) dataset \cite{Wang_2021_visionlan}, which includes the Weakly Occluded Scene Text (WOST) dataset and the Heavily Occluded Scene Text (HOST) dataset. The two datasets pose a unique challenge in that a portion of text is obscured, therefore accurately recognizing the text requires the model to have certain text reasoning capability. As shown in Tab. \ref{tab:ost}, all IGTR models exhibit superiority over the compared methods, i.e., SVTR-B \cite{duijcai2022svtr}, LPV-B \cite{ijcai2023LPV} and MAERec \cite{jiang2023revisiting}, indicating the advantages of instruction-guided learning in augmenting the capture of linguistic association. Interestingly, IGTR-PR-PT and IGTR-AR-PT achieve much higher accuracy compared to the IGTR-based competitors, especially in the more challenging HOST dataset. This observation suggests that the amalgamation of increased training data with instructional guidance effectively models the linguistic context of the text, and leading to enhanced performance.

In Tab.~\ref{tab:chinese}, we also give the results on CTR \cite{chen2021benchmarking}, a Chinese benchmark. IGTR models still outperform existing methods by clear margins. Further accuracy gains are observed when targeted strengthening (TS) is employed. The improvements come from the better recognition of rarely appearing and morphologically similar characters. In addition, we observe that data augmentation is particularly useful for the CTR benchmark. Adding the augmentation can further improve the accuracy up to 3.80\% for IGTR-PR-TS and 4.57\% for IGTR-AR-TS. We argue that, on one hand, the image modality constitutes a minority in terms of quantity when trained following the instruction-guided paradigm. On the other hand, the CTR dataset is not big enough for Chinese with thousands of character categories. Data augmentation can mitigate the two issues especially for difficult text, resulting in more prominent improvements on challenging subsets such as Scene and Handwriting. To sum, the experiments also validate the superiority of IGTR and its great cross-language generalization ability. Note that for Chinese the model size increases moderately due to the incorporation of a larger character vocabulary compared to English.

\subsection{Distinction with existing pre-training methods}
Previous methods introduced several general pre-training techniques into STR. They can be roughly categorized into three types: contrastive learning-based methods like SeqCLR \cite{AberdamLTASMMP21_SeqCLR} and PerSec \cite{Liu_2022_PerSec}, mask auto-encoder-based methods like MAERec \cite{jiang2023revisiting}, and methods like DiG \cite{YangLLWZLTB22_DiG} that combines both techniques. We select the open-sourced SeqCLR, MAERec, and DiG from the three types and compare our IGTR-PR-PT and IGTR-AR-PT with them. For the three existing methods, IGTR architecture is first pre-trained on Union14M-U \cite{jiang2023revisiting}, an unlabeled real-world dataset containing approximately 10 million samples. Afterwards, the results in Tab. \ref{tab:pretrained} are obtained by fine-tuning on Union14M-L training set. Since IGTR-PR-PT and IGTR-AR-PT are both pre-trained on automatically rendered synthetic datasets, all the compared methods do not require manual annotation and their fine-tuning step is the same. The two training protocols constitute a fair comparison among the methods.

As depicted in Tab. \ref{tab:pretrained}, models pre-trained based on DiG get slightly better results than those pre-trained based on SeqCLR and MAERec, highlighting the merit of combining both contrastive learning-based and mask auto-encoder-based pre-training. Moreover, all three pre-training methods positively contribute to training better recognition models, where performance gains are consistently observed when comparing the raw IGTR-PR and IGTR-AR results in \ref{tab:sota_u14m}. The results confirm the effectiveness of using pre-training techniques. Moreover, IGTR-PR-PT outperforms DiG-based IGTR-PR by 0.69\% in Common and by 2.01\% in Union14M-L. Similarly, IGTR-AR-PT surpasses DiG-based IGTR-AR by 0.44\% in Common and by 0.91\% in Union14M-L. The results indicate that our attribute prediction-based pre-training provides more beneficial knowledge for text recognition tasks compared to the three pre-training methods, even when only trained on synthetic datasets. We posit that this superiority arises from the diversity of our instructions, which leads to a more comprehensive pre-training regime and consequently better model training. This reaffirms the significance of understanding diverse character attributes.

\begin{figure}
 \centering
\includegraphics[width=0.48\textwidth]{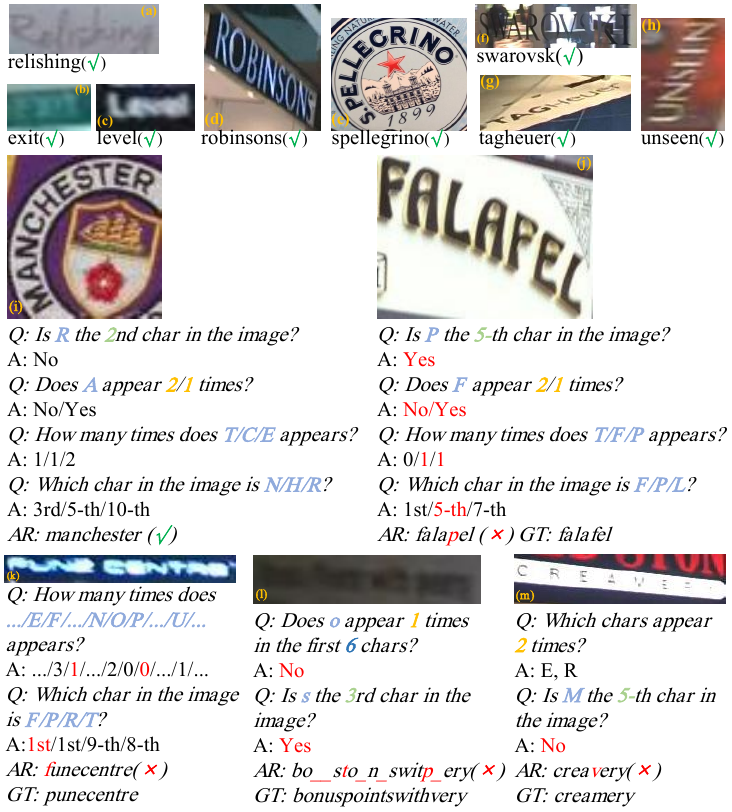} 
 \caption{Success and failure cases. All results are obtained with IGTR-AR trained on Union14M-L training set. Wrong answers are marked in \textcolor{red}{red}.}
 \label{fig:figcase}
\end{figure}

\subsection{Qualitative Analyses}

To qualitatively understand IGTR, we depict some success and failure cases in Fig. \ref{fig:figcase}. Cases (a)-(i) are all IGTR correctly recognized images, which are rather challenging and many previous methods have failed on them. We illustrate case (i) in detail by asking some questions and presenting the model's responses, which are all correct. Note that for current STR models, the model itself could not answer these questions without consulting the recognized text. This example shows the merit of our instruction-guided learning. It well understands character attributes thus leads to correct text reasoning. Nevertheless, there also are a few failure cases. For cases (j)-(m), answers to all the asked questions are logically consistent although misrecognized. For example, case (j) has made exactly the same mistake of recognizing the fifth \emph{F} as \emph{P}. Furthermore, cases (k), (l) and (m) correspond to severely blurred, low-contrast, and overexposed text, respectively. They are quite challenging even for humans. However, a majority of characters are still correctly inferred for these examples, demonstrating that our instruction-guided learning has evoked powerful character understanding capability. 

\begin{figure}
 \centering
\includegraphics[width=0.48\textwidth]{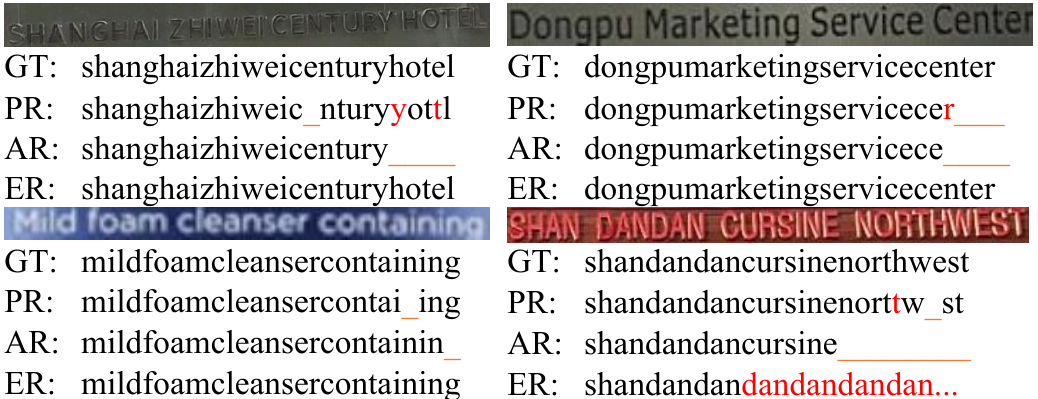} 
 \caption{Results of three recognition pipelines on long text. Incorrect predictions are marked in \textcolor{red}{red} and missed predictions are denoted by \textcolor{red}{\_}. PR, AR, and ER are IGTR-PR, IGTR-AR, and IGTR-ER, respectively.}
 \label{fig:figlong}
\end{figure}

To assess the extrapolating-based recognition, in Fig. \ref{fig:figlong} we give four examples of long text recognition, which is also a typical STR challenge. The results show that IGTR-ER can recognize text exceeding 25 characters. This is because IGTR-ER does not encode the absolute position embedding. Therefore it can overcome this character limit. As can be seen, IGTR-ER correctly recognizes three challenging long text cases that IGTR-PR and IGTR-AP both failed. Nevertheless, the right-bottom case indicates that IGTR-ER may trigger the error of repeated recognition if the same sub-string appears multiple times, which constitutes a practical issue to address.

We also inspect IGTR models from the attention map perspective. As shown in the bottom of Fig. \ref{fig:fig1}, IGTR-AR's attention maps $Attn=Sum(Attn_1,Attn_2,\ldots,Attn_h)$, where $Attn_i$ (Eq. \ref{equ:mha}) generated by \emph{Question to Image Attention and FFN} in Fig. \ref{fig:IGTR_overview} is visualized, can provide a qualitative explanation of its superior performance. When performing attribute prediction, the attention maps show that IGTR-AR can accurately pinpoint and focus on characters emphasized in the questions. This implies that IGTR-AR effectively correlates the instructions and the image content through the proposed CFMM module. The image patterns are well understood thus enabling smooth execution of the related tasks. As a contrast, previous methods are localized less satisfactory and struggle with recognizing rarely seen text image patterns (as shown in the upper of Fig. \ref{fig:fig1}).

\subsection{More Discussion}
For modern text recognizers, recent research focus mostly lies in enhancing recognition performance through various means such as extracting discriminative visual features, incorporating linguistic information, etc. We delve into the distinctions between IGTR and several popular STR models, including SVTR \cite{duijcai2022svtr}, NRTR \cite{Sheng2019nrtr}, ABINet \cite{fang2021abinet}, and PARSeq \cite{BautistaA22PARSeq}, from this perspective. Specifically, SVTR is a purely visual model, where the superior recognition performance stems primarily from enhanced visual features. NRTR introduced linguistic clues by leveraging previously decoded characters, which matches the AR-based recognition instruction in Tab. \ref{tab:rec_instruction}. PARSeq introduced a learning pattern where a subset of characters is provided, and the model deduces the remaining characters within the text. This learning pattern is encompassed by specific conditions and questions in IGTR, e.g., the first and fourth conditions in Tab. \ref{tab:instruction}, the \emph{Search status} question, etc. ABINet employs a language model-based character double-check mechanism, akin to the Re-identification instruction in Tab. \ref{tab:rec_instruction}. On the other hand, IGTR adopts SVTR-B as its visual backbone, and devises a diverse set of instructions to model character-central attributes comprehensively. By doing so, the linguistic information is fully integrated, while the STR models just discussed, from the linguistic modeling point of view, can be viewed as only utilizing distinct subsets of IGTR instructions. This explains why IGTR achieves superior recognition performance. Moreover, the instruction-guided training in IGTR also enjoys great flexibility, targeted strengthened models can be easily constructed by simply adjusting the instructions sampling, which is difficult to achieve by existing STR models. This showcases another advantage of IGTR in terms of adaptability.

\section{Conclusion}

In this paper, we have presented IGTR, an instruction-guided multi-modal paradigm to build accurate, fast and lightweight STR models. To gain these attractive properties, we have extended traditional question-answer instruction to $\left \langle condition,question,answer\right \rangle$ triplet, which not only increases the diversity and richness of the instructions but also allows efficient parallel learning. We have developed a lightweight instruction encoder, a cross-modal feature fusion module and a multi-task answer head. They together generate a novel architecture with sufficient cross-modal feature interactions, contributing to IGTR models that present leading accuracy on various English and Chinese benchmarks. Notably, IGTR enjoys the flexibility of instruction sampling, by using different text recognition instructions, IGTR models following different recognition pipelines are constructed and achieve different trade-offs between accuracy and inference speed. Moreover, by adjusting the sampling of attribute prediction instructions, a targeted strengthened IGTR model can be built, which better recognizes rarely appearing and morphologically similar characters. It is worth noting that the size of IGTR models is uniformly 24.1M, and they consume 3.98 ms-10.3 ms on average using one NVIDIA 1080Ti GPU when inferring a text instance. Both are appealing properties in resource-constrained applications.

This paper has made a meaningful attempt in developing small and efficient multi-modal models dedicated to a specific task, i.e., STR. There also are many works ahead. One is long text recognition \cite{iccv2023lister}. Long text is prevalent in real-world applications, as current tools may not correctly separate multiple words in a line and text recognizers should take them as a whole. Our preliminary attempt has observed the problem of repeated recognition when directly using IGTR-ER. How to tackle this problem is worthy of further study. Another critical issue is \emph{small} OCR foundation model, which is capable of accurately and efficiently processing multiple OCR-related tasks like text recognition, document understanding, etc. This research would benefit a wide array of applications related to OCR. We believe that similar initiatives in other domains \cite{wu2024building,zhu2023minigpt4,ren2024grounding} can shed light on us.

\bibliographystyle{IEEEtran}
\bibliography{egbib}

\end{document}